\documentclass[Afour,sageh,times]{format/sagej}

\usepackage{moreverb,url}
\usepackage{natbib}
\usepackage[usenames,dvipsnames,svgnames,table]{xcolor}
\usepackage{wrapfig}
\usepackage{times}
\usepackage{multicol}
\usepackage{multirow}
\usepackage[bookmarks=true, draft]{hyperref}
\usepackage{graphicx}
\usepackage{amsmath,amsthm,amssymb}
\usepackage{pgf,tikz}
\usepackage{mathrsfs}
\usetikzlibrary{arrows}
\usepackage[linesnumbered,ruled,vlined]{algorithm2e}
\usepackage{epstopdf}
\usepackage{overpic}
\usepackage{xcolor}
\usepackage{xargs} 
\usepackage{verbatim} 
\usepackage[textsize=footnotesize]{todonotes}
\usepackage{moresize}
\usepackage{enumerate}
\usepackage{placeins} 
\usepackage{marvosym}
\usepackage{tabularx}
\usepackage{IEEEtrantools}

\usepackage{soul}
\setstcolor{blue}

\SetCommentSty{mycommfont}

\usetikzlibrary{
    arrows,
    arrows.meta,
    shapes,
    shapes.geometric,
    decorations.markings}
\tikzset{every node/.style={thick,shape=circle,draw=black,fill=blue!15,text=black,minimum size=1.7em,inner sep=0.5}}
\tikzset{every picture/.style=thick}
\tikzset{>={Latex[length=2mm]}}
\colorlet{customLightRed}{red!65}     

\definecolor{Gray}{gray}{0.9}
\SetKwRepeat{Do}{do}{while}%

\newcommand{\super}{{\tt Super4PCS}}

\newcommand{\apc}{{\tt APC}}
\newcommand{\icp}{{\tt ICP}}

\newcommand{\cnn}{{\tt CNN}}
\newcommand{\rgbd}{{\tt RGB-D}}

\newcommand{\dof}{{\tt DoF}}

\newcommand{\object}{O}
\newcommand{\model}{M}
\newcommand{\trans}{T}

\DeclareMathOperator*{\argmax}{arg\,max}

{\end{list}}

\begin{document}

\runninghead{Mitash, Boularias and Bekris}

\title{Physics-based Scene-level Reasoning for Object Pose Estimation in Clutter}

\author{Chaitanya Mitash, Abdeslam Boularias and Kostas Bekris}

\corrauth{Chaitanya Mitash, Abdeslam Boularias, Kostas Bekris,\\
Computer Science Department, Rutgers University,\\
Piscataway, NJ 08854}

\email{\{cm1074, ab1544, kb572\}@rutgers.edu}

\begin{abstract}
This paper focuses on vision-based pose estimation for multiple rigid
objects placed in clutter, especially in cases involving occlusions
and objects resting on each other. Progress has been achieved recently
in object recognition given advancements in deep learning.
Nevertheless, such tools typically require a large amount of training
data and significant manual effort to label objects. This limits their
applicability in robotics, where solutions must scale to a large
number of objects and variety of conditions. Moreover, the
combinatorial nature of the scenes that could arise from the placement
of multiple objects is hard to capture in the training dataset. Thus,
the learned models might not produce the desired level of precision
required for tasks, such as robotic manipulation. This work proposes
an autonomous process for pose estimation that spans from data
generation to scene-level reasoning and self-learning. In
particular, the proposed framework first generates a labeled dataset
for training a Convolutional Neural Network (\cnn) for object
detection in clutter. These detections are used to guide a scene-level
optimization process, which considers the interactions between the
different objects present in the clutter to output pose estimates of
high precision. Furthermore, confident estimates are used to label
online real images from multiple views and re-train the process in a self-learning pipeline. Experimental results indicate that
this process is quickly able to identify in cluttered scenes
physically-consistent object poses that are more precise than the ones
found by reasoning over individual instances of objects. Furthermore,
the quality of pose estimates increases over time given the
self-learning process.

\end{abstract}

\keywords{Object Detection, 6D Pose Estimation, Robot Perception,
  Convolutional Neural Networks, Monte Carlo Tree Search, Lifelong Learning}
\maketitle

\section{Introduction}
\label{sec:intro}
A critical capability of a robot is to be able to identify the 6-DoF
poses of objects in their surroundings so as to be able to manipulate
them. Many environments, however, contain cluttered scenes, where
objects are placed in complex arrangements, and can only be partially
observed from the robot's viewpoint due to occlusions. An example of
such a setup exists in current day warehouses, where robots are being
deployed for tasks such as picking from bins, packing and sorting.
Recently, deep learning methods, such as those employing Convolutional
Neural Networks (\cnn s), have become popular for object detection
(\cite{ren2015faster, redmon2016you}) and pose estimation
(\cite{kehl2017ssd, xiang2017posecnn}), outperforming alternatives in
object recognition benchmarks. These desirable results are typically
obtained by training \cnn s using datasets that involve a very large
number of labeled images. However, these datasets need to be
collected in a way that captures the intricacies of the environment
the robot is deployed in, such as lighting conditions, occlusions and self-occlusions, in clutter.

\begin{figure}[t]
\centering
\includegraphics[width=\linewidth, keepaspectratio]{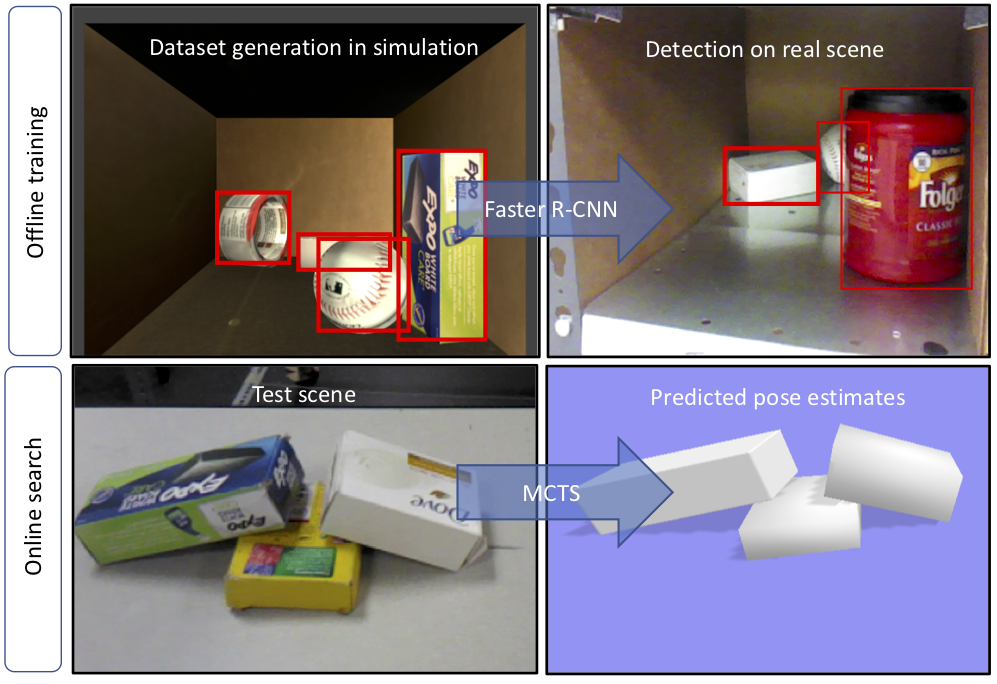}
\caption{(top) A physically-realistic dataset is generated and is used
to train a \cnn\ for object detection. (bottom) In the online phase,
pose estimation is performed via a Monte Carlo tree search process,
which performs scene-level reasoning to output physically-realistic pose estimates of higher-accuracy.}
\label{fig:into}
\end{figure}

The recent Amazon Picking Challenge (\apc) (\cite{Correll:2016aa}) has
reinforced this realization and has led into the development of
datasets specifically for the detection of objects inside cluttered,
shelving units (\cite{singh2014bigbird, rennie2016dataset,
Princeton}). These datasets are created either with human annotation
or by incrementally placing one object in the scene and using
foreground masking. An increasingly popular approach to avoid manual
labeling is to use synthetic datasets generated by rendering 3D CAD
models of objects with different viewpoints. Synthetic datasets have
been used to train \cnn s for object detection
(\cite{peng2015learning}) and viewpoint estimation
(\cite{su2015render}). One major challenge in using synthetic data is
the inherent difference between virtual training examples and real
testing data. For this reason, there is considerable interest in
studying the impact of texture, lighting, and shape to address this
disparity (\cite{SunVirtual}). Another issue with synthetic images
generated from rendering engines is that they display objects in poses
that are not necessarily physically realistic. Moreover, occlusions
are usually treated in a rather na\"ive manner, i.e., by applying
cropping, or pasting rectangular patches, which again results in
unrealistic scenes (\cite{peng2015learning, su2015render,
movshovitz2016useful}).

This motivates the development of a synthetic dataset that could
capture the known parameters of the environment and generate data
accordingly. It should also be able to avoid overfitting to the
unknown parameters. The first key idea presented in this work is the
use of a physics engine in synthetic dataset generation pipeline. The
physics engine defines environmental constraints on object placement,
which naturally capture in the training set, the distribution of
object poses that can realistically appear during testing.
Furthermore, a physics engine is a very convenient tool to
parameterize the unknown scene features, such as illumination. Randomization over such parameters is very effective in avoiding
overfitting to synthetic textures of objects.

Even after detecting all the objects in a given image using a trained {\tt CNN}, the problem of estimating the 6-DoF poses of the objects in the 3D workspace involves geometric reasoning regarding the position
and orientation of the detected objects. Very recently end-to-end
learning for 6-DoF pose estimation was proposed by \cite{kehl2017ssd}
and \cite{xiang2017posecnn}. These methods predict the approximate
6-DoF poses and are often followed by an online local optimization
process in the form of \icp ( \cite{icp}). Other solutions that have
been developed, use a Convolutional Neural Network (\cnn) for object
segmentation (\cite{Princeton, hernandez2016team}) followed by a {\tt
3D} model alignment step using point cloud registration techniques
(\cite{mellado2014super, icp}). The quality of the pose estimate,
however, can still suffer due to over-reliance on the learned models.

The second key observation of this work is to treat individual-object predictions with some level of uncertainty and perform a global, scene-level optimization process that takes object interactions into account. This information arises from physical properties, such as respecting gravity and friction as well as the requirement that objects do not penetrate one another. Through this physical reasoning, which is achieved by incorporating physics simulations, the resulting pose estimates for the objects are of improved accuracy and by default consistent. In this way, they can be directly used in the context of manipulation planning framework.

Once the system has access to an object detector and a pose estimation process, it could already be deployed for the desired application. However, as the system performs its task, it gets access to data in the operation domain which it did not have access to initially. This data could be very useful in further improving the performance of the system. Nevertheless, this data is not labelled. This motivates the need for automatically labeling real images and adding them to the existing synthetic dataset. 

Overall, the current work has two contributions. Algorithmically, this work proposes a Monte Carlo Tree Search (MCTS) based optimization process for scene reasoning with physics-based priors. The MCTS based algorithm searches over the cartesian product of individual object pose candidates to find the optimal scene hypothesis with respect to a score defined in terms of similarity of rendered hypothesized scenes with the input data. The search performs constrained local optimization for each of these candidate object poses via physics correction and {\tt ICP}. This helps in pruning a large search space and thus quickly achieving accurate pose estimates. Secondly, this work provides a complete pipeline for 6-DoF pose estimation of objects placed in clutter. The main components of the proposed pipeline include:
\begin{itemize}

\item {\bf A physics simulation tool}, which uses scene information,
such as the placement of a resting surface (e.g., tabletop, shelf,
etc.), object models and camera calibration to set up an environment
for generating training data. The tool performs physics simulation to
place objects at realistic configurations and renders images of scenes
to automatically generate a synthetic dataset to train an object
detector. This tool exploits the known environmental constraints and
randomizes the unknown parameters to generate a dataset, which
captures to a good extent, the properties of clutter.

\item {\bf A self-learning process}, which employs a robotic
manipulator to autonomously collect multi-view images of real scenes
and to label them automatically using the object detector trained with
the above physics-based simulation tool. The key insight behind this
system is the fact that the robot can often find a good viewing angle
that allows the detector to accurately label the object and estimate
its pose. The object's predicted pose is then used to label images of
the same scene taken from more difficult and occluded views. The
transformations between different views are known because they are
obtained by moving the robotic manipulator.

\end{itemize}

The proposed pipeline is evaluated over 4 challenging datasets, namely, the {\tt Shelf\&Tote} dataset (\cite{Princeton}), {\tt Linemod} (\cite{hinterstoisser2012model}), {\tt Linemod-Occluded} (\cite{brachmann2014learning}) and the {\tt Extended Rutgers RGBD dataset} which was collected by the authors and labelled to test the applicability of the physics-based scene-level reasoning process. 

On the {\tt Shelf\&Tote} dataset, the proposed pipeline, which bootstraps pose estimation with a synthetic dataset outperforms state-of-the-art systems that have access to labelled real images. {\tt Extended Rutgers RGBD dataset} was collected to reflect different levels of physical dependencies between objects. Evaluation over this dataset shows that the {\tt MCTS} based reasoning could quickly identify physically-realistic accurate poses for complex setups where approaches that consider individual object instances fail to provide a good solution. Finally, the entire pipeline was evaluated on {\tt Linemod} and {\tt Linemod-Occluded} according to a recently published benchmark on pose estimation (\cite{hodan2018bop}) and the proposed approach outperforms several state-of-the-art techniques on this task.

This paper is an integration of two conference articles by the same authors into a complete framework for object pose estimation (\cite{mitash2017self} and \cite{mitash2018improving}). It expands upon the technical details provided in the aforementioned articles and provides additional examples as well as evaluations. In particular, it contributes a comprehensive process for object pose estimation, which starts with the generation of training data in physics-based simulation, followed by the steps of congruent set matching to generate object pose hypothesis, pose clustering to reduce the cardinality of the hypothesis set, and a scene-level optimization process to get accurate pose estimates. It also discusses how the obtained pose estimates can be used in a self-learning process to reduce the domain gap that might exist between the simulated training data and real test scenes. Dataset and code for the entire pipeline is publicly shared. ({\small \url{http://www.physimpose.com}})

\section{Related Work}
\label{sec:related}
This section discusses the different methodologies for object pose
estimation and their relation to the current work.

\subsection{Local Point Descriptors}

One popular approach to pose estimation is to match feature points
between textured 3D models and images (\cite{Lowe:1999aa,
  Rothganger:2006aa, Collet:2011aa}). This requires, however, textured
objects and good lighting conditions, which has motivated instead the
use of range data. Some range-based techniques compute correspondences
between local point descriptors on the observed scene and on the
object {\tt CAD} models. Once correspondences are established, robust
detectors like generalized Hough transform
(\cite{ballard1981generalizing}) or {\tt RANSAC}
(\cite{fischler1981random}) are used to compute the rigid transform
that is consistent with the majority of correspondences. Several local
descriptors are available (\cite{aldoma2012tutorial}), such as
signature of histograms of orientations ({\tt SHOT})
(\cite{tombari2010unique}), fast point feature histogram ({\tt FPFH})
(\cite{rusu2009fast}) and Spin Images (\cite{johnson1999using}). There
has also been work on improving the efficiency of {\tt RANSAC} and
Hough transform (\cite{tombari2010object,
  papazov2010efficient}). Feature-based approaches can be extended to
multi-view object recognition (\cite{Pillai:2015aa}) and pose
estimation (\cite{Erkent:2016aa}) so as to increase accuracy relative
to single frame estimates. This family of methods depends on local
surface information, which is sensitive to the resolution and quality
of sensor and model data. The features are often parametrized by the
area of influence, which is not trivial to decide. The smaller area could
lead to less discriminative features between different surfaces on the
object, while a larger area could result in sensitivity to occlusion
and noise.

\subsection{Oriented Point Pair Features}

One proposed way to counter these limitations is to use \emph{oriented
  point pair features} (\cite{drost2010model}) so as to create a
global object model in the form of a map that stores the model points
that exhibit each feature. This map can then be used to match the
features in the scene and to get the object pose through a fast voting
scheme. This idea was later extended to incorporate color
(\cite{choi20123d}), geometric edge information (\cite{drost20123d})
and visibility context (\cite{kim20113d}).  Recently, point pair
features were used for segmenting the scene into several clusters,
where each cluster generates a separate pose hypothesis
(\cite{birdal2015point}). The votes are weighted based on the
probability of visibility of model points. Recent work
(\cite{hinterstoisser2016going}) uses a sampling strategy for scene
points by reasoning about the size of the object model. The approach
modifies the voting scheme to accommodate sensor noise by also voting
in the neighboring bins. Point pair features have been criticized in
some occasions for performance loss in the presence of background
clutter, sensor noise and also due to their quadratic computational
complexity.

\subsection{Template Matching and Coordinate Regression}

Another category of methods for pose estimation is based on \emph{template matching}, such as {\tt LINEMOD} (\cite{hinterstoisser2012model}) and variants like \cite{hodavn2015detection}. This method is based on viewpoint sampling around a 3D CAD model and building templates for each viewpoint based on color gradient and surface normals, which are later matched to compute object pose. {\tt GPU}-based implementations help to speed-up computation (\cite{Cao:2016aa}). Other popular approaches (\cite{brachmann2014learning, tejani2014latent, krull2015learning}) are based on learning to predict 3D object coordinates in the local model frame. A recent effort (\cite{michel2017global}) performs geometric validation on these predictions using a conditional random field. Performance of these approaches can be highly dependent on the predictions of three-dimensional object coordinates from the random field, which are not trivial to train. Template matching approaches, on the other hand, often fail to reason about occlusions.

\subsection{Deep Learning}

The success of deep learning on problems related to object detection and semantic segmentation (\cite{ren2015faster, shelhamer2016fully})
has motivated their use for aspects of pose estimation or for the development of direct pipelines for pose estimation (\cite{xiang2017posecnn}). Inspired by the applicability of \cnn s for descriptor learning of \rgbd\ views (\cite{Wohlhart:2015aa}), recent work (\cite{Kehl:2016aa}) has demonstrated deep learning of descriptive features from local \rgbd\ patches used to create 6D pose hypotheses. Similarly, \cnn s have been used to detect semantic keypoints to estimate the 6 \dof\ pose consistent with the keypoints (\cite{Pavlakos:2017aa}). Deep learning has also been integrated in a principled way with a global search for the discovery of 3 \dof\ poses of multiple objects (\cite{narayanan2016discriminatively}). There are also other data-driven approaches for identifying features for object recognition (\cite{Bo:2014aa}). The success of these approaches often depends on representative labelled training data. Also, these methods are often followed by a local optimization process, such as {\tt ICP}, which is not always sufficient for fixing the errors and ambiguities in predictions. The current work leverages the success of deep learning in the task of object segmentation. It considers, however, the uncertainties in individual object predictions to guide a global optimization process to estimate poses.

\subsection{Registration Methods}

Many recent pose estimation techniques (\cite{Princeton, hernandez2016team, mitash2018improving}) integrate \cnn s for segmentation with pointset registration techniques (\cite{mellado2014super}). Popular local registration approaches are Iterative Closest Points (\icp) (\cite{icp}) and its variants (\cite{Rusinkiewicz:2001aa, Mitra:2004aa, Segal:2009aa, Bouazix:2013aa, Srivatsan:2017aa}), which typically require a good initialization. Otherwise, registration requires finding the best aligning rigid transform over the {\tt 6-DOF} space of all possible
transforms, which are uniquely determined by 3 pairs of (non-degenerate) corresponding points. A popular strategy is to invoke {\tt RANSAC} to find aligned triplets of point pairs (\cite{Irani:1996aa}) but suffers from a frequently observable worst case $O(n^3)$ complexity in the number $n$ of data samples, which has motivated many extensions (\cite{Gelfand:2005aa, Cheng:2013aa}). The {\tt 4PCS} algorithm (\cite{aiger20084}) achieved $O(n^2)$ output-sensitive complexity using four congruent point basis instead of three. This method was extended to \super\ (\cite{mellado2014super}), which achieves $O(n)$ output-sensitive complexity. The accuracy of these methods, however, highly depends on the predictions from the object detector.

\begin{figure*}[ht]
\centering
\includegraphics[width=\textwidth, height=7.5cm, keepaspectratio]{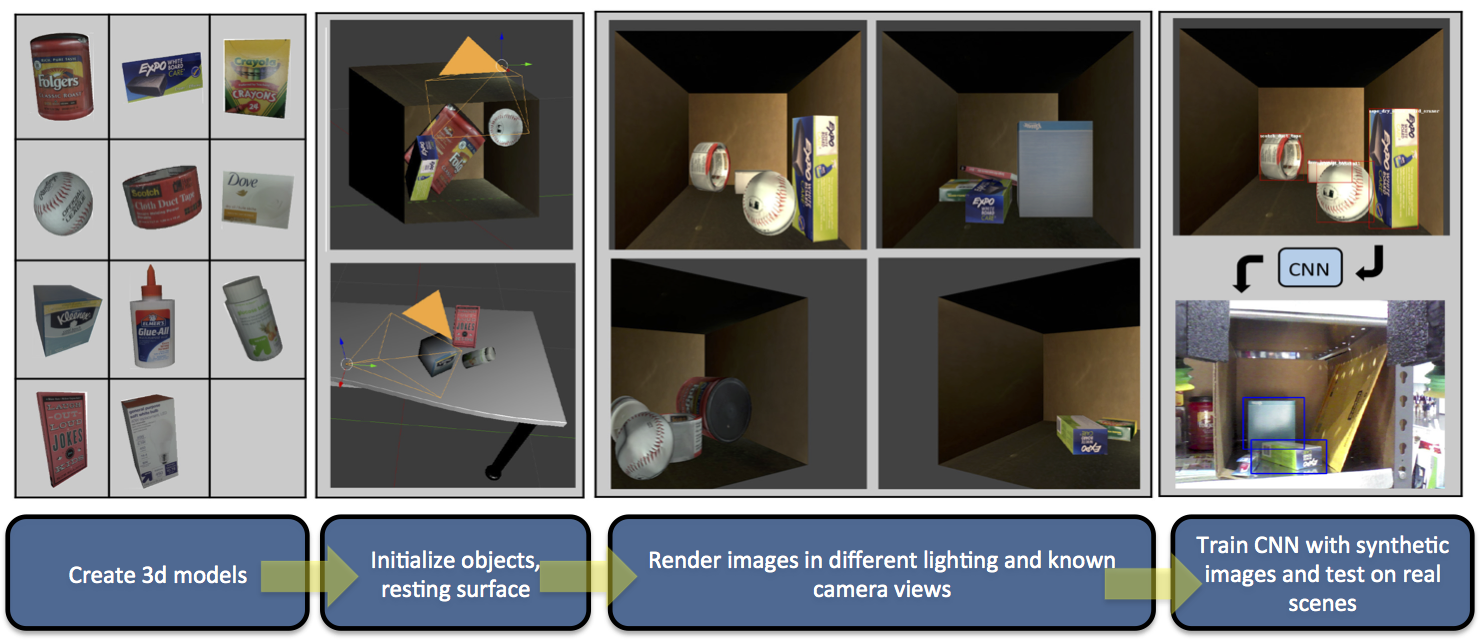}
\caption{Pipeline for physics aware simulation: The 3D CAD models are
  generated and loaded in a calibrated environment on the simulator. A
  subset of the objects is chosen for generating a scene. Objects are physically simulated until they settle on the resting surface under the effect of gravity. The scenes are rendered from known camera poses. Perspective projection is used to compute 2D bounding boxes
  for each object. The labeled scenes are used to train a Faster-RCNN object detector (\cite{ren2015faster}), which is tested on a real-world setup.}
\label{fig:physim}
\end{figure*}

\subsection{Simultaneous localization and mapping (SLAM)}
SLAM (\cite{durrant2006simultaneous, thrun2005probabilistic}) is a popular problem in robotics which deals with constructing the map of an unknown environment with a sensor mounted on a robot while simultaneously keeping track of the location of the robot in the world frame. Recently, a popular strategy in this field is object-based SLAM (\cite{salas2013slam++, mccormac2018fusion++}), which performs object pose estimation and tracking using a depth sensor and uses the relative configurations of the objects to reason about the location of the camera. Several such approaches also make use of synthetic datasets with simulated camera trajectories to learn semantic information for indoor scenes (\cite{mccormac2017scenenet}). There have also been efforts (\cite{stein2018genesis}) at bridging the domain gap between these synthetic scenes and sensor-acquired images for semantic labeling via image translation techniques such as CycleGAN (\cite{zhu2017unpaired}). Although these work have a notion of using synthetic dataset for learning semantic scene segmentation and of applying scene-level constraints, this problem space is quite different from that of object pose estimation in cluttered scenarios.

\subsection{Global Scene-Level Reasoning}

A popular approach to resolve conflicts arising from local reasoning
is to generate object pose candidates and perform a Hypothesis
Verification ({\tt HV}) step (\cite{aldoma2012global,
  aldoma2013multimodal, akizuki2016physical}). The hypotheses
generation in most cases occurs using a variant of {\tt RANSAC}
(\cite{RANSAC, mellado2014super}). One of this method's drawbacks is
that the generated hypotheses might already be conflicted due to
errors in object segmentation and thus performing an optimization over
this might not be very useful. A recently proposed method reasons
globally about the hypotheses generation process
(\cite{michel2017global}). Nevertheless, this requires explicit
training for pixel-wise predictions. Another approach to counter these
drawbacks corresponds to an exhaustive but informed search to find the
best scene hypotheses over a discrete set of object placement
configurations (\cite{narayanan2016discriminatively,
  narayanan2016discriminatively}). A tree search formulation as
described above was defined to effectively search in {\tt 3-DOF}
space. It is not easy, however, to apply the method for {\tt 6-DOF}
pose estimation due to scalability and resolution issues.

This work shows that by training an object detector with an autonomous
clutter-aware process, it is possible to generate a set of object
candidate poses by a fast global point cloud registration method,
which only has local geometric conflicts. Generating candidate poses
in this manner and then applying a search process, which constrains
each object expansion to other object placements leads to significant
improvements in the final pose estimation results.

\section{Problem Setup}
\label{sec:problem}
This work considers the problem of estimating the 6D poses of $N$ known objects $\{ \object_1, \ldots, \object_N \}$ in a scene,
captured by an {\tt RGB-D} camera. The knowledge of the following elements is assumed:
\begin{itemize}
	\item geometric models are given as textured triangular meshes $\{ \model_1, \ldots, \model_N \}$ of all the objects that are present in the scene. Mass of objects are kept as constant across all objects and friction as well as linear and angular damping coefficients for objects are set to maximum within the simulator.
	\item triangular mesh and pose $\trans_{rs}$ for the resting surface of the objects, such as a shelf or a table in a global reference frame,
	\item the intrinsic and extrinsic parameters $K, \trans_{cam}$ for the camera.
\end{itemize}
The estimated poses are returned as a set of rigid-body transformations $\{ \trans_1, \ldots, \trans_N \}$, where each $T_i = (t_i, R_i)$
captures the translation $t_i \in R^3$ and rotation $R_i \in SO(3)$ of object model $\model_i$ in a globally defined reference frame.

\section{Approach}
\label{sec:approach}
This section presents the proposed approach for object recognition and
pose estimation. It first describes how a dataset of labeled images
could be generated autonomously to train a convolutional neural
network (\cnn) for object detection. It then outlines, how the
detection output of \cnn s is used in a search process to obtain 6-DoF
pose estimates of multiple objects present in the scene. Finally, it
describes a self-learning pipeline that uses the pose
estimation output to label real images from multiple views and
re-train the detector to improve its performance.

\subsection{{\bf Generating training dataset}}

The first component of the proposed framework physically simulates scenes containing target objects and generates images of the corresponding scenes using the parameters of a known camera. This is used to generate a synthetic dataset for training a \cnn -based object detector. The pipeline for this process is depicted in Figure~\ref{fig:physim}.

The dataset generation process mimics a real-world setup involving a sensing system for robotic manipulation, where a camera is mounted on a robotic arm. The robot is placed in front of a surface for object placement (resting surface), such as a shelf-bin or table-top, which contains the objects. In such a setup, forward kinematics can be used to provide the 6-DoF pose $\trans_{cam}$ of the camera. Furthermore, a camera calibration process provides the intrinsic parameters of the camera K. The pose of the resting surface $\trans_{rs}$ relative to the robot is determined by a {\tt RANSAC}-based estimation process (\cite{RANSAC}). For instance, for the shelf depicted in Figure~\ref{fig:physim}, such a pose estimation process was implemented by computing the edges and planes on the retrieved depth data and matching them against the known geometry of the shelf.

Given the above information as input, the method aims to render and automatically label several images in simulation as discussed in Algorithm~\ref{alg:alg1}.

\begin{algorithm}[h]
\caption{{\sc physim\_cnn}$( \trans_{cam}, \trans_{rs}, K, M_{1:N})$ }
\tcp{$\trans_{cam}$: set of camera poses for rendering}
\tcp{$\trans_{rs}$: pose of the resting surface}
\tcp{K: intrinsic camera parameters}
\tcp{$M_{1:N}$: mesh models for all N objects}
\label{alg:alg1}
dataset $\gets \emptyset$\;
\While{ ( $|dataset| <$ desired size ) }
{
    O $\gets$ a random subset of objects from $M_{1:N}$\;
    $\trans^{O}_{init} \gets $ {\sc initial\_random\_poses}( O )\;
    \tcp{random intial pose within a specified domain is assigned to each object in $O$}
    $\trans^{O}_{final} \gets$ {\sc physics\_sim}($\trans^{O}_{init}, \trans_{rs}$, O)\;
    \tcp{physics simulation is performed to obtain the final, physically consistent object pose for objects in $O$}
    Light $\gets$ {\sc pick\_lighting\_conditions}()\;
    \ForEach { $(view \in \trans_{cam})$ }
    {
        image $\gets$ {\sc render}( $\trans^{O}_{final}$, view, K, Light)\;
	   \{ labels, bboxs \} $\gets$ {\sc project}($\trans^O_{final}$, view)\;
       \tcp{set of object poses $\trans^O_{final}$ is used to generate bounding-boxes in all views}
        dataset $\gets$ dataset $\cup$ (image, labels, bboxs);
    }
}
Train {\sc faster-rcnn} with the generated dataset\;
\end{algorithm}

The algorithm simulates a scene by first selecting randomly a set of objects O from the list of available object models $M_{1:N}$ (line 3). The initial pose of an object is provided by function {\sc initial\_random\_poses} (line 4), which samples uniformly at random along the x and y-axis from the range $(\frac{-dim_i}{2}, \frac{dim_i}{2})$, where $dim_i$ is the dimension of the resting surface along the $i^{th}$ axis. The initial position along the z-axis is fixed and can be adjusted to either simulate dropping or placing. The initial orientation is sampled appropriately in {\tt SO(3)}. Then, function {\sc physics\_sim} is called (line 5), which physically simulates the objects and allows them to fall due to gravity, bounce, and collide with each other as well as with the resting surface. Any inter-penetrations among objects or with the surface are treated by the physics engine. The final poses of the objects ${\bf P}_{final}^{O}$, when they stabilize, resemble real-world poses. Gravity, friction coefficients, and mass parameters are set at similar values globally and damping parameters are set to the maximum to promote fast stabilization.

The environment lighting and point light sources are varied with respect to location, intensity, and color for each rendering (line 6). Simulating various indoor lighting sources helps to avoid over-fitting to a specific texture, which makes the training set more robust to different testing scenarios. Once lighting conditions are chosen, the simulated scene is rendered from multiple views using the pre-defined camera poses (line 6). The rendering function {\sc render} requires the set of stabilized object poses $\trans_{final}^{O}$, the camera viewpoint as well as the selected lighting conditions and intrinsic camera parameters (line 7). Finally, perspective projection is applied to obtain 2D bounding box labels for each object in the scene with function {\sc project} (line 8). The overlapping portion of the bounding boxes for the object that is further away from the camera is pruned.

The generated synthetic dataset is used to train an object detector based on {\tt Faster-RCNN} (\cite{ren2015faster}), which utilizes a deep {\tt VGG} network architecture (\cite{simonyan2014very}). The dataset generation module has been implemented using the {\tt Blender} API, which internally uses the {\tt Bullet} physics engine and has been publicly shared (\small \url{https://github.com/cmitash/physim-dataset-generator}).

A critical requirement for learning with synthetic data as discussed above is the need for modeling the domain in the simulation. The precision with which the geometry and texture of the objects and support surface need to be modeled depends on the set of objects to be detected. If the objects have very different geometries, a noisy modeling of the shape using surface reconstruction technique like KinectFusion (\cite{izadi2011kinectfusion}) is good enough for the recognition task, such as in the {\tt Linemod} dataset (\cite{hinterstoisser2012model}). If there are multiple objects with similar geometry, accuracy in modeling the texture and color is more critical to achieving a good performance, for example in the {\tt Shelf\&Tote} dataset (\cite{Princeton}). Other physical properties like mass and friction coefficients of objects have been kept as constant over all objects for the scope of this work while object material properties and parameters corresponding to the illumination of the environment have been randomized within a wide domain.

\begin{figure*}[h!]
\centering \includegraphics[width=\textwidth]{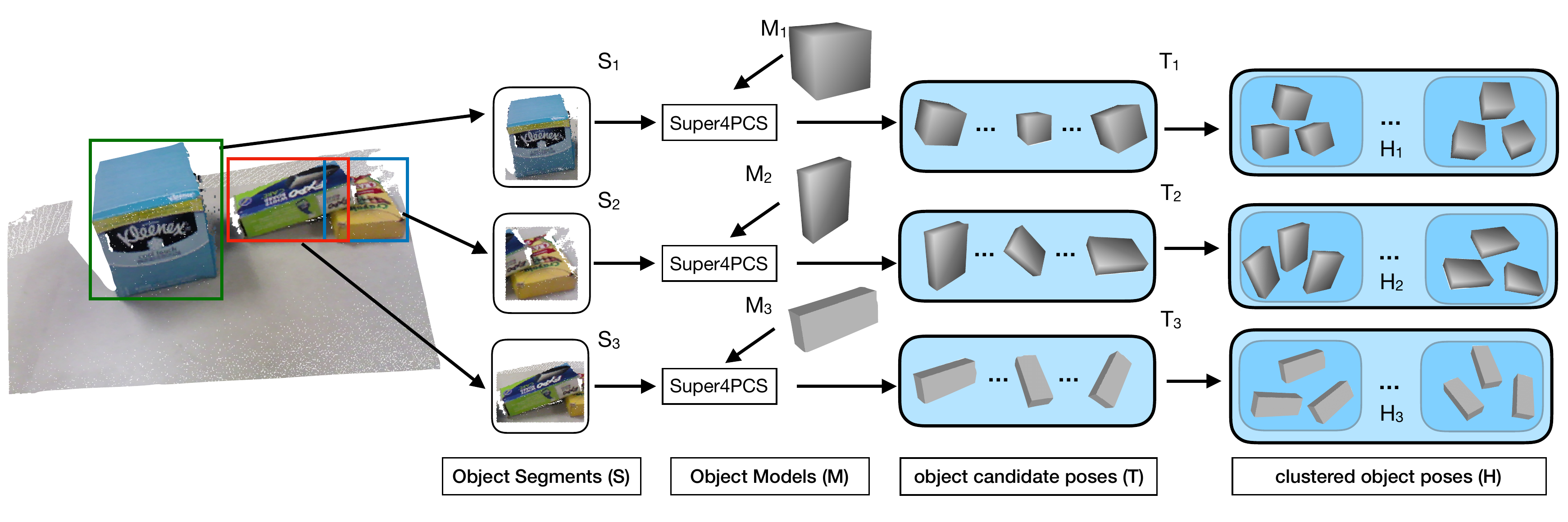}
\caption{The image describes the process of hypotheses generation for
objects present in the scene. The process starts with extracting
object segments S$_{1:3}$ using Faster-RCNN, followed by a congruent
set matching process to compute a set of possible model
transformations (T$_{1:3}$) that correspond to the respective
segments. These transformations are then clustered to produce object-
specific hypotheses sets (H$_{1:3}$).}
\label{fig:hypogen}
\end{figure*}

\subsection{{\bf Pose Estimation}}

The next component of the system is a method for 6-DoF pose
estimation. It proceeds by:

\begin{enumerate}
\item generating a set of pose hypotheses based on the detections from
the previously trained detector for each object present in the scene,
and
\item searching efficiently over the set of joint hypotheses for
the most globally consistent solution.
\end{enumerate}

Global consistency is quantitatively evaluated by a score
function. The score function measures the similarity between the
actual observed depth image and a rendering of the objects in
simulation using their hypothesized poses. The hypothesized poses are
adapted during the search process, so as to correspond to poses where
the objects are placed in a physically realistic and stable
configuration according to a physics engine that simulates rigid
object dynamics.

\subsubsection{{\bf Hypothesis Generation:}}

Some of the desired properties for a set of 6D pose hypotheses are the
following:
\begin{itemize}
\item informed and diverse enough such that the optimal solution is
either already contained in the set or a close enough hypothesis
exists so that a local optimization process can fine-tune it and
return a good result;
\item limited in size, as evaluating the dependencies among the
hypotheses set for different objects can lead to a combinatorial
explosion of possible joint poses and significantly impact the
computational efficiency;
\item does not require extensive training.
\end{itemize}

This work considers all of these properties while generating the
hypothesis set. The pseudocode for hypothesis generation is presented
in Algorithm~\ref{alg:alg3}.

\begin{algorithm}[h]
\caption{{\sc gen\_hypothesis}({\tt RGB,depth,$M_{1:N}$} ) }
\label{alg:alg3}
\tcp{Given an RGB-D image and a set of object models $M_{1:N}$, {\sc gen\_hypothesis} generates pose candidates $h_O$ for each object $O$.}
$H \gets \{h_O = \emptyset, \forall O \in M_{1:N}\}$\;
\ForEach{ object O in the scene }{
    $bbox_O \gets$ {\sc rcnn\_detect}( {\tt RGB}, O)\;
    \tcp{bounding box is detected for object O using the trained Faster-RCNN detector.}
    $P_O \gets$ {\sc get\_3dpoints}( bbox$_O$, {\tt depth})\;
    \tcp{3d point cloud of object O is extracted from the depth image according to bbox}
    $T_O \gets$ {\sc congruent\_set\_matching}(M$_O$, P$_O$)\;
    \tcp{a set of pose candidates is generated as illustrated in Fig.~\ref{fig:cSet}}
    $\{cluster_{tr}, center_{tr}\} \gets$ {\sc kmeans}$_{tr}(T_O$)\;
    \tcp{candidate poses are clustered according to their translation vectors using the KMeans algorithm.}
    \ForEach{ cluster C in cluster$_{tr}$ }{
        $\{cluster_{rot}, center_{rot}\} \gets$ {\sc k-kmeans}$_{tr}(C)$\;
    	\tcp{candidate poses within cluster C are further clustered according to their euler angles using Kernel-KMeans.}
        $h_O \gets h_O \cup (center_{tr}, center_{rot})$\;
    }
    $H \gets H \cup h_O$\;
}
return H;
\end{algorithm}

The detector trained with the autonomous training process proposed in
the previous section is used to extract bounding-box ({\tt bbox$_O$})
for each object O in the scene. This, in turn, gives a segment P$_O$
of the 3D point cloud. Segment P$_O$ is a subset of the point cloud of
the scene and contains points from the visible part of the object
O. Segment P$_O$ frequently contains some points from nearby objects
because the bounding box does not perfectly match the shape of the
object.

\begin{figure}[t]
\centering \includegraphics[width=\linewidth]{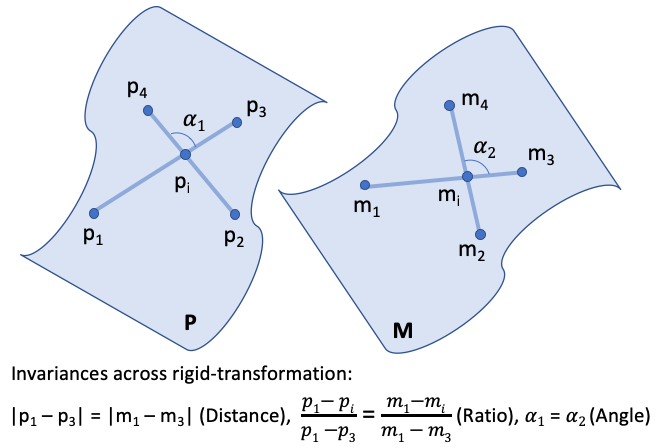}
\caption{Figure depicts the {\tt congruent-set-matching} process which finds sets of 4-points on the scene and on the object model that are congruent under rigid transformation. All the point-pairs on the model M with distances similar to $\mid p_1$ - $p_3\mid$ and $\mid p_2$ - $p_4\mid$ on the sampled base, can be found in linear time using an efficient technique as described in \cite{mellado2014super}. Then 4-point congruent sets are found by evaluating these point-pairs based on other invariances such as ratios and angles.}
\label{fig:cSet}
\end{figure}

The received point set P$_O$ is then matched to the object model M$_O$ in the subroutine {\sc congruent\_set\_matching} in Algorithm~\ref{alg:alg3} to generate pose candidates for the object. This module, inspired by the {\tt Super4PCS} (\cite{mellado2014super}) algorithm iteratively samples a set of 4 co-planar points from P$_O$ called the base and finds sets of 4-points on the model which are congruent under rigid transformation, to the base. Each pair of congruent sets gives a pose hypothesis. The matching process is depicted in Fig.~\ref{fig:cSet}. The fact that the distances, angles, and ratios of the intersection of line segments are maintained over a rigid transform is used to come up with an efficient linear time algorithm for finding the congruent sets. The time complexity of this process is $O(n + m + k)$, where n is the number of points on the sampled object model, m is the number of point-pairs on the model which are at the same distance as a point-pair on the sampled base and k is the number of the congruent sets found corresponding to the sampled base. As opposed to RANSAC (\cite{RANSAC}) which has a time complexity of $O(n^3)$ for matching a set of 3 points to all triplets of points on the model, this process better exploits the geometric constraints from rigid transformations and efficiently produces a relatively small set of pose candidates.

Nevertheless, {\tt Super4PCS} evaluates each of these transformations to find the one that achieves the best alignment according to the {\tt LCP} (Largest Common Pointset) metric. This returned transformation, however, is not necessarily the optimal object pose as the point cloud segment extracted via the detection process could include parts of other objects or due to lack of visible surface might not be informative enough to compute the correct solution. This motivates the consideration of other possible transformations for the objects, which can be evaluated in terms of scene-level consistency.

Thus, the proposed process retains a set of possible transformations \trans$_O$ computed using congruent set matching within a given time budget t$_o$. It is interesting to consider the quality of the hypotheses set returned by the above process by measuring the error between the returned pose hypotheses and the ground truth. For this purpose, a validation dataset containing 90 object poses was used. Specifically, in each hypothesis set, the pose hypothesis that has the minimum error in terms of rotation is selected as well as the one with the minimum translation error. The mean errors for these candidates over the dataset are shown in Table.~\ref{table:metricEval}. The results positively indicate the presence of hypotheses close to the true solution. Specifically, the candidate with the minimum rotation error seems almost perfect in the rotation and not very far even with respect to translation. Nevertheless, this hypothesis set contained approximately 20,000 elements. It is intractable to evaluate scene-level dependencies for that many hypotheses per object as the combined hypotheses set over multiple objects grows exponentially in size.

\begin{table*}[ht]
\small\sf\centering
\caption{Evaluating the quality of the hypotheses set returned by Super4CPS with respect to different metrics.\label{table:metricEval}}
\begin{tabular}{lll}
\toprule
Metric for selection&Mean Rotation error&Mean Translation error (\%)\\
\midrule
{[All hypotheses] max. LCP score} & 11.16$^{\circ}$ & 1.5cm\\
{[All hypotheses] min. rotation error from ground truth} & 2.11$^{\circ}$ & 2.2cm\\
{[All hypotheses] min. translation error from ground truth} & 16.33$^{\circ}$ & 0.4cm\\
{[Clustered hypotheses] min. rotation error from ground truth} & 5.67$^{\circ}$ & 2.5cm\\
{[Clustered hypotheses] min. translation error from ground truth} & 20.95$^{\circ}$ & 1.7cm\\
\bottomrule
\end{tabular}\\[10pt]
\end{table*}

\subsubsection{{\bf Clustering of Hypotheses:}}

To reduce the cardinality of the hypotheses sets returned by the
subroutine {\sc congruent\_set\_matching} in Algorithm~\ref{alg:alg3},
this work proposes to cluster the 6D poses in each set T$_O$, given a
distance metric. Computing distances between object poses, which are
defined in SE(3), in a computationally efficient manner is not trivial
(\cite{zhang2007c}). This challenge is further complicated if one
would like to consider the symmetry of the geometric models, so that
two different poses that result in the same occupied volume given the
object's symmetry would get a distance of zero.

To address this issue, a two-level hierarchical clustering approach is
followed. The first level involves computing clusters of the pose set
in the space of translations (i.e., the clustering occurs in
$\mathbb{R}^3$ by using the Euclidean distance and ignoring the object
orientations) using a K-Means process (\cite{arthur2007k}) to get a
smaller set of cluster representatives {\tt cluster$_{tr}$}. In the
second level, the poses that are assigned to the same clusters are
further clustered based on a distance computed in the SO(3) space that
is specific to the object model, i.e., by considering only the
orientation of the corresponding pose. The second clustering step uses
a {\it kernel} K-Means approach (\cite{dhillon2004kernel}), where the
cluster representative is found by minimizing the sum of kernel
distances to every other point in the cluster. This process can be
computationally expensive but returns cluster centers that nicely
represent the accuracy of the hypotheses set. By using this clustering
method, the size of the hypotheses set can be reduced down from 20,000
rigid transforms in T$_O$ to 25 object pose hypotheses in h$_O$ for
each object in the scene. The two bottom rows of
Table \ref{table:metricEval} evaluate the quality of the cluster
representatives in the hypotheses set. This evaluation indicates that
the clustering process returns hypotheses as cluster representatives
that are still close to the true solution. In this way, it provides an
effective way of reducing the size of the hypotheses set without
sacrificing its diversity.

\subsubsection{{\bf Search:}}

Once the hypotheses set is built for each object in the scene, the
task reduces to finding the object poses that lie in the physically
consistent neighborhood of the pose candidates and best explain the
overall observed scene. In particular, given:
\begin{itemize}
\item the observed depth image $I_D$, 
\item the number of objects in the scene {\tt N},
\item a set of 3D mesh models for these objects {\tt M}$_{1:N}$,
\item and the sets of 6D transformation hypotheses for the objects h$_{1:N}$ (output of Algorithm~\ref{alg:alg3}),
\end{itemize}
the problem is to search in the hypotheses sets for an N-tuple of
poses T$_{1:N}$ so that $T_i \in {\it f}(h_i)$, i.e., one pose per
object. The set $T_{1:N}$ should maximize a global score computed by
comparing the observed depth image with the rendered image
R(T$_{1:N}$) of object models placed at the corresponding poses
T$_{1:N}$. Here, {\it f} is the constrained local optimization of the
object pose {\tt h$_i$} based on physical consistency with respect to
the other objects in the scene and also the fact that the same points
in the scene point cloud cannot be explained by multiple objects
simultaneously. Then, the global optimization score is defined as:

\begin{eqnarray*}
C(I_D, T_{1:N}) = \sum_{p \in P}Sim(R(T_{1:N})[p], I_D[p])
\end{eqnarray*}

\noindent where $p$ is a pixel (i,j) of a depth image, $R(T_{1:N})[p]$
is the depth of pixel $p$ in the rendered depth image, $I_D[p]$ is the
depth of pixel $p$ in the observed depth image, $P = \{p \mid
R(T_{1:N})[p] \neq 0 \text{ or } I_D[p] \neq 0)\}$ and

\begin{eqnarray*}
Sim(R(T_{1:N})[p], I_D[p]) =
\begin{cases}
1, \text{ if } \mid R(T)[p] - I_D[p] \mid < \epsilon \cr
0, \text{otherwise}
\end{cases}
\end{eqnarray*}

\begin{algorithm}[ht]
\tcp{$s_d$: state at depth d (pose assignment for first d objects)}
\tcp{$(M_{d+1}, T_{d+1})$: mesh model and pose hypothesis for the $(d+1)^{th}$ object}
\tcp{$P_{d+1}$: point cloud segment for $(d+1)^{th}$ object}
\caption{{\sc expand}$( s_d, (M_{d+1}, T_{d+1}), P_{d+1} )$ }
\label{alg:alg4}
\If{d $=$ N}{
	return NULL\; \tcp{maximum depth of tree is reached}
} 
\ForEach{ $(M_O, T_O)$ $\in s_d$ }{
	$P_{d+1} \gets P_{d+1}$ - {\sc points\_explained}($P_{d+1}, M_O, T_O$)\;
} \tcp{remove points from $P_{d+1}$ already assigned to an object $M_O$ at $T_O$ in $s_d$}
$T_{d+1} \gets ${\sc trimmed\_icp}$((M_{d+1}, T_{d+1}), P_{d+1})$\; \tcp{pose is locally refined using trimmed-ICP}
$T_{d+1} \gets ${\sc physics\_sim}$((M_{d+1}, T_{d+1}), s_d)$\; \tcp{pose is locally refined based on physics simulation}
$s_{d+1} \gets s_d \cup (M_{d+1}, T_{d+1})$\;
return $s_{d+1}$;
\end{algorithm}

\SetKwProg{Fn}{Function}{}{}
\begin{algorithm}[h!]
\tcp{$M_{1:N}$: mesh models for all N objects}
\tcp{$P_{1:N}$: point cloud segments for all N objects}
\tcp{$h_{1:N}$: pose candidate sets for all N objects}
\caption{{\sc search} }
\label{alg:alg5}
\Fn{{\sc mcts} ($M_{1:N}, P_{1:N}, h_{1:N}$)}{
	S $\gets \emptyset$\;
	$L_{1:K} \gets$ {\sc get\_dependency}$(P_{1:N})$\;
	\tcp{$L_{1:K}$ is a partition of N objects into K subsets where objects belonging to different subsets are physically independent of each other}
	\ForEach {$L \in L_{1:K}$}{
		$s_0 \gets \emptyset$\;
		$best\_render\_score \gets 0$\;
		$best\_state \gets s_0$\;
		\While{$search\_time < t_{th}$}{
		\tcp{$t_{th}$ is a pre-defined time budget.}
			$s_i \gets$ {\sc select}$(s_0, M_{1:N}, P_{1:N}, h_{1:N})$\;
			\tcp{$s_i$ is the next state to be expanded based on UCB}
			$\{s_N, R\} \gets$ {\sc random\_policy}$(s_i, M_{1:N}, P_{1:N}, h_{1:N})$\;
			\tcp{$s_N$ is the state obtained by randomly selecting poses for all unplaced objects, i.e. not in $s_i$. R is the rendered score for $s_N$}
			\If{$R > best\_render\_score$} {
				$best\_render\_score \gets R$\;
				$best\_state \gets s_R$\;
			}
			{\sc backup\_reward}($s_i, R$)\;
			\tcp{$R$ is used to update estimated costs of all states s along the path from $s_i$ to the root node.}
		}
	$S \gets T \cup best\_state$\;
	}
	return S\; \tcp{S is a set of object poses for N objects}
}
\end{algorithm}

\begin{algorithm}[h]
\caption{{\sc search modules} }
\label{alg:alg_components}
\Fn{{\sc select} ($s, M_{1:N}, P_{1:N}, h_{1:N}$)}{
\tcp{$s$: state of the search tree}
\tcp{$M_{1:N}$: mesh models for all N objects}
\tcp{$P_{1:N}$: point cloud segments for all N objects}
\tcp{$h_{1:N}$: pose candidate sets for all N objects}
	\While{$depth(s) < N$}{
		\eIf{$s$ has unexpanded child}{
			d $\gets$ $depth(s)$\;
			T$_{d+1}$ $\gets$ {\sc next\_pose\_hypothesis}(h$_{d}$)\;
			\tcp{$T_{d+1}$ is the next pose candidate for (d+1)th object that has not already been expanded for state s.}
			return {\sc expand}($s, (M_{d+1}, T_{d+1}), P_{d+1}$)\;
			\tcp{appends the pose $T_{d+1}$ to state s}
		}{
			Return best child $s$ according to UCB equation~\ref{uct}\;
		}
	}
	return s\;
}
\BlankLine
\BlankLine
\Fn{{\sc random\_policy} ($s, M_{1:N}, P_{1:N}, h_{1:N}$)}{
\tcp{$s$: state of the search tree}
\tcp{$M_{1:N}$: mesh models for all N objects}
\tcp{$P_{1:N}$: point cloud segments for all N objects}
\tcp{$h_{1:N}$: pose candidate sets for all N objects}
	\While{$depth(s) < N$}{
		$d \gets depth(s)$\;
		$T_{d+1} \gets$ {\sc get\_random\_hypothesis}(h$_{d+1}$)\;
		\tcp{$T_{d+1}$ is a random pose assigned to the $(d+1)^{th}$ object.}
		$s \gets$ {\sc expand}($s, (M_{d+1}, T_{d+1}), P_{d+1}$)\;
		\tcp{appends the pose $T_{d+1}$ to state s.}
	}
	return \{s, render(s)\}\;
}
\BlankLine
\Fn{{\sc backup\_reward} ($s, R$)}{
\tcp{$s$: state of the search tree}
\tcp{$R$: render score for the state s}
	\While{s $\neq$ NULL}{
		n(s) $\gets$ n(s) + 1\;
		h(s) $\gets$ h(s) + R\;
		\tcp{number of expansions and estimated cost of state s is updated}
		s $\gets$ parent(s)\;
	}
}
\end{algorithm}

\noindent for a predefined precision threshold $\epsilon$. Therefore, score {\tt C} counts the number of non-zero pixels $p$ that have a similar depth in the observed image $I_D$ and in the rendered image {\tt R} within an $\epsilon$ threshold. So, overall the objective is to find:
$$T^*_{1:N} = \arg\max_{T_{1:N} \in {\it f}(h_{1:N})} C( I_D,
R(T_{1:N})).$$

At this point, a combinatorial optimization problem arises so as to identify $T^*_{1:N}$, which is approached with a tree search process. A state in the search-tree corresponds to a subset of objects in the scene and their corresponding poses. The root state s$_0$ is a null assignment of poses. A state $s_d$ at depth {\tt d} is a placement of $d$ objects at specific poses selected from the hypotheses sets, i.e., {\tt s}$_d = \{{(M_i, T_i),i=1:d}$\} where $T_i$ is the pose chosen for object $M_i$, which is assigned to a tree depth $i$. The goal of the tree search is to find a state at depth {\tt N}, which contains a pose assignment for all objects in the scene and maximizes the above-mentioned rendering score. Alg.~\ref{alg:alg4} describes the expansion of a state in the tree search process towards this objective.

The {\sc expand} routine takes as input the state {\tt s}$_d$ at tree depth $d$, the point cloud segment corresponding to the next object to be placed, $P_{d+1}$, and the pose hypothesis $T_{d+1}$ for the next object to be placed $M_{d+1}$. Lines 3-4 of the algorithm iterate over the objects already placed in state $s_d$ and remove points explained by these object placements from the point cloud segment of the next object to be placed. This step helps in achieving much better segmentation, which is utilized by the local optimization step of Trimmed ICP (\cite{chetverikov2002trimmed}) in line 5. The poses of objects in state s$_d$ physically constrain the pose of the new object to be placed. For this reason, a rigid body physics simulation is performed in line 6. The physics simulation is initialized by inserting the new object into the scene at pose $T_{d+1}$, while the previously inserted objects in the current search branch are stationary in the poses $T_{1:d}$. A physics engine is used to ensure that the newly placed object attains a physically realistic configuration (stable and no penetration) with respect to other objects and the table under the effect of gravity. After a fixed number of simulation steps, the new pose $T_{d+1}$ of the object is appended to the previous state to get the successor state s$_{d+1}$.

The above primitive is used to search over the tree of possible object
poses. The objective is to exploit the contextual ordering of object
placements given information from physics and occlusion. This does not
allow to define an additive rendering score over the search depth as
in previous work (\cite{narayanan2016discriminatively}), which demands the object
placement to not occlude any part of the already placed
objects. Instead, this work proposes to use a heuristic search
approach based on Monte Carlo Tree Search utilizing the {\it Upper
Confidence Bound} formulation (\cite{kocsis2006bandit}) to trade off
exploration and exploitation in the expansion process. The pseudocode
for the search is presented in Alg.~\ref{alg:alg5} and
Alg.~\ref{alg:alg_components}.

\begin{figure*}[ht]
\centering
\includegraphics[width=\textwidth, height=7.5cm, keepaspectratio]{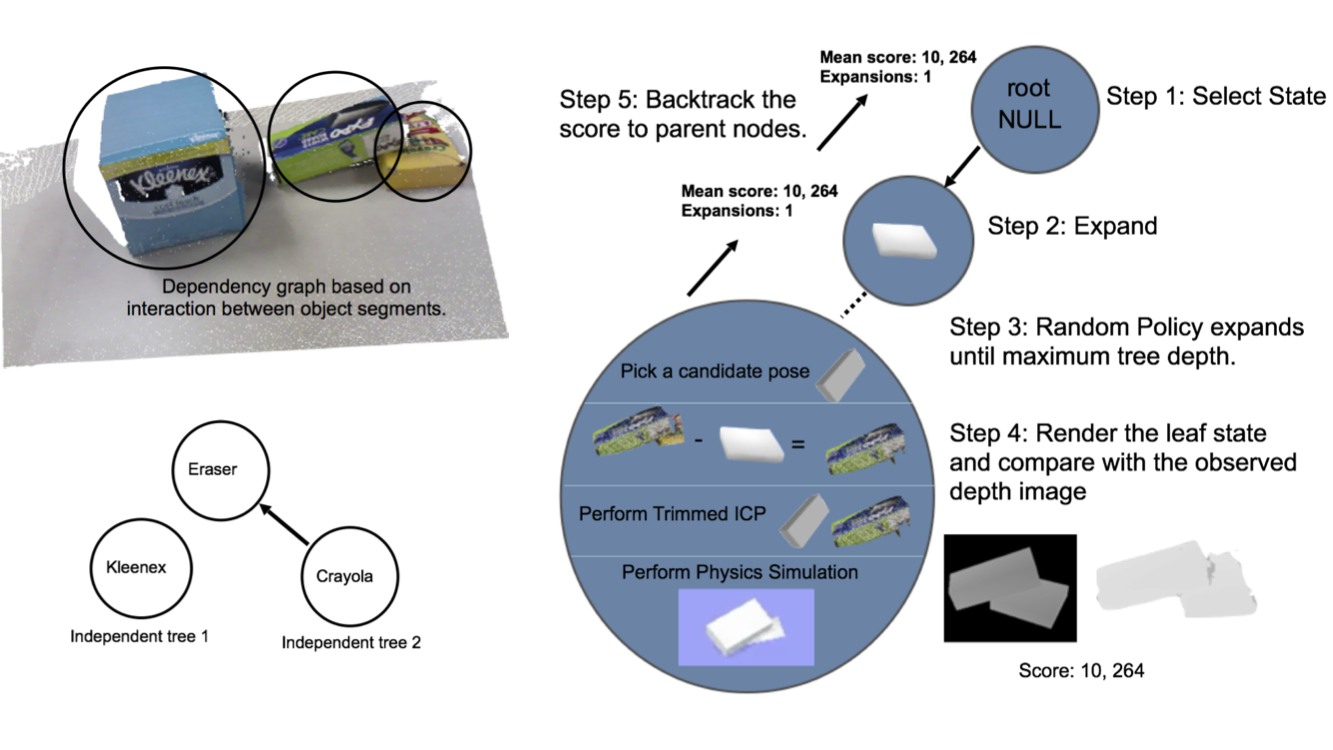}
\caption{(Left) Image shows how a dependency graph is built based on interactions between the object segments. The Kleenex object is placed separately and does not need to be evaluated in the same tree like the others, whereas the placement of expo depends on the placement of crayola. (Right) The image shows one iteration of the Monte Carlo Tree Search process. The next state to expand is selected based on the previously computed score for the states. The selected state is then evaluated by executing a random policy which keeps expanding state until all objects in the current tree are placed. Finally, a rendering of this completely reconstructed scene is compared to the observed depth image to compute a score for the state.}
\label{fig:expand}
\end{figure*}

To effectively utilize the constrained expansion of states, an order of object placements needs to be considered. This information is
encoded in a {\tt dependency graph}, which is a directed acyclic graph
that provides a partial ordering of object placements but also encodes
the interdependency of objects. An example of a dependency graph
structure is presented in Fig.\ref{fig:expand}. The vertices of the
{\tt dependency graph} correspond to the objects in the observed
scene. Simple rules are established to compute this graph based on the
detected segments {\tt P$_{1:N}$} for objects O$_{1:N}$.
\begin{itemize}
\item A directed edge connects object $O_i$ to object $O_j$ if the x-y projection of $P_i$ in the world frame intersects with the x-y projection of $P_j$ and the z-coordinate (negative gravity direction) of the centroid for $P_j$ is greater than that of $P_i$.
\item A directed edge connects object $O_i$ to object $O_j$ if the detected bounding-box of $O_i$ intersects with that of $O_j$ and the z-coordinate of the centroid of $P_j$ in camera frame (normal to the camera) is greater than that of $P_i$.
\end{itemize}

The information regarding the independence of objects helps to significantly speed up the search as the independent objects are then evaluated in different search trees and prevent exponential growth of the tree. This results in {\tt K} ordered list of objects, {\tt L}$_{1:K}$ coming from the module {\sc get\_dependency} of Alg.~\ref{alg:alg_components}, each of which is passed to an independent tree search process for pose computation. The {\tt MCTS} proceeds by selecting the first unexpanded node starting from the root state. The selection of the next state to be expanded takes place based on a reward associated with each state. The reward is the mean of the rendering score received at any leaf node in the state's subtree along with a penalty based on the number of times this subtree has been expanded relative to its parent. This is the {\tt Upper Confidence Bound (UCB)} formulation (\cite{kocsis2006bandit}). Formally, given a state s of the search tree, the next state to be expanded is selected as,

\begin{equation}
s = \argmax_{s' \in succ(s)} \frac{h(s')}{n(s')} + \alpha \sqrt{\frac{2log(n(s))}{n(s')}}\;
\label{uct}
\end{equation}
where h(s) is the estimated score for state s, n(s) is the number of times the subtree rooted at the state s has been expanded and $\alpha$ is the parameter that controls the trade-off between exploration and exploitation in the search process. The selected state is then expanded by using a {\sc random\_policy}, which in this case is picking a random object pose hypothesis for each of the succeeding objects while performing the constrained local optimization at each step. The output of this policy is the final rendering score of the generated scene hypotheses. This reward is then backpropagated in the step {\sc backup\_reward} to all preceding nodes. Thus, the search is guided to the part of the tree, which gets a good rendering score but also explores other portions, which have not been expanded enough (controlled by the parameter $\alpha$). Figure~\ref{fig:expand} visualizes these steps of the {\tt MCTS} pipeline.

\section{Self-learning}

\begin{figure*}[ht]
\centering
\includegraphics[width=0.8\textwidth, height=7cm]{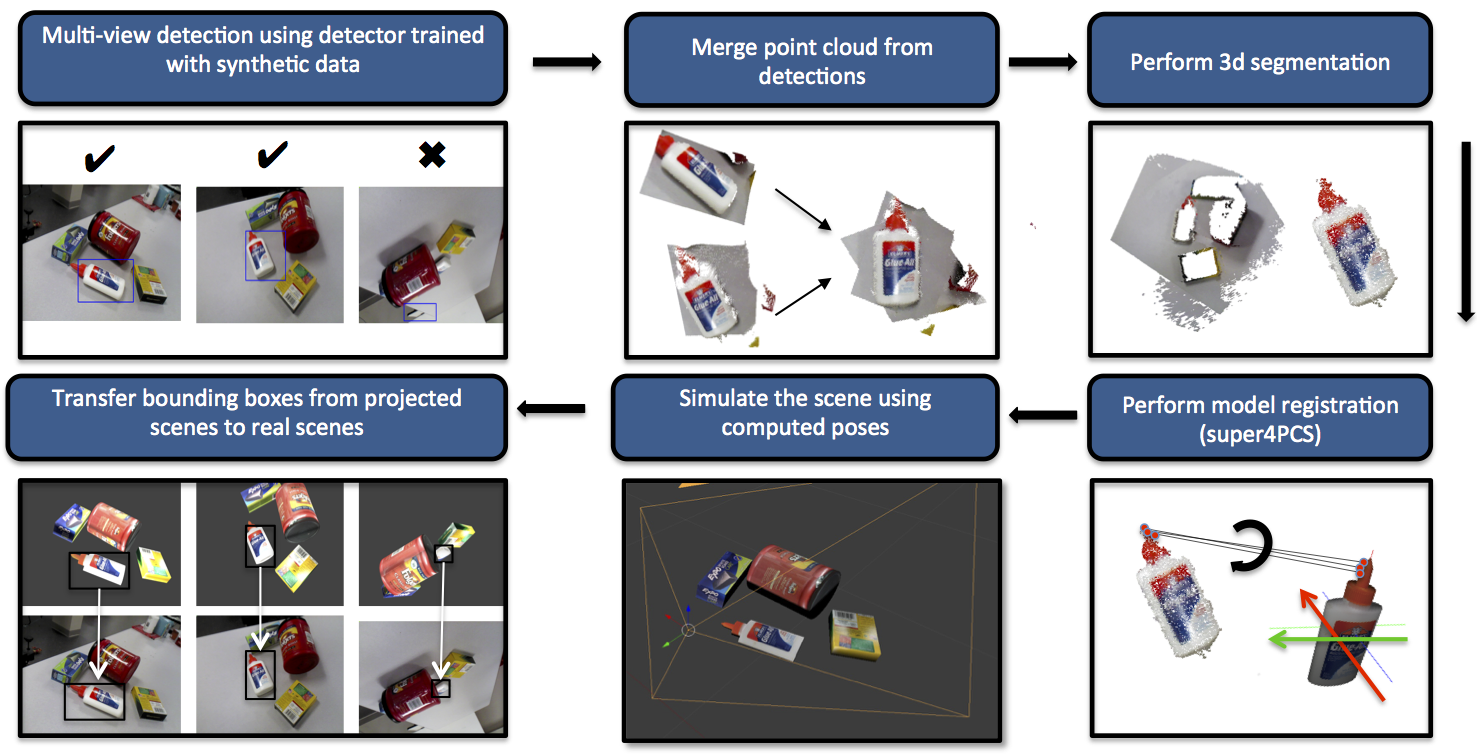}
\caption{Automatic self-labeling pipeline: The detector trained with simulated data is used to detect objects from multiple views. The point cloud aggregated from successful detections undergoes 3D segmentation. Then, Super4PCS (\cite{mellado2014super}) is used to estimate the 6D pose of the object in the world frame. The computed poses with high confidence are projected back to the views to obtain precise labels over real images.}
\label{fig:selflearn}
\end{figure*}

Given access to an object detector and a pose estimation process
trained with the physics-based simulator, the self-learning pipeline
labels real-world images with a robust multi-view pose
estimation. This is based on the idea that the detector performs well
on some views, while might be imprecise or fail in other
views. Aggregating 3D data over the confident detections and with
access to the knowledge of the environment, a 3D segment can be
extracted for each object instance in the scene. This process,
combined with the fact that 3D models of objects are available, makes
it highly likely to estimate correct 6-DoF poses of objects given
enough views and search time. The results of pose estimation are then
projected back to the multiple views and used to label real
images. These examples are very effective to reduce the confusion in
the classifier for novel views. The process also autonomously
reconfigures the scene using manipulation actions to apply the
labeling process iteratively over time in different scenes, thus
generating a labeled dataset which is used to re-train the object
detector. The pipeline of the process is presented in
Fig.~\ref{fig:selflearn} and the pseudocode is provided in
Algo.~\ref{alg:alg6}.

A robotic arm is used to move the sensor to different pre-defined
camera configurations $\trans_{cam}$ and capture RGB and depth images
of the scene (lines 2-3). The PRACSYS motion planning library
(\cite{kimmel2012pracsys}, \cite{Littlefield:2015aa}) was used to
control the robot in the accompanying implementation.

\begin{algorithm}[ht]
\caption{{\sc self-learn}$( dataset, \trans_{cam}, M_{1:N}$) }
\label{alg:alg6}
\tcp{dataset: synthetic training dataset}
\tcp{$\trans_{cam}$: set of camera poses to collect images}
\tcp{$M_{1:N}$: mesh models for all N objects}

\While{$|dataset| <$ desired size}
{
  \ForEach{ view $\in \trans_{cam}$ }
  {
    \{RGB$_{view}$, D$_{view}$\} $\gets$ {\sc capture}( view )\;
  } 
  \tcp{RGB-D images are collected my moving the camera to all views in $T_{cam}$}
  \ForEach{object O in the scene}
  {
    $Cloud_O = \emptyset$\;
    \ForEach{ view $\in \trans_{cam}$}
    {
      bbox $\gets$ {\sc sim\_detect}( RGB$_{view}$ )\;
      \tcp{bounding box is detected for object $O$ in image $RGB_{view}$}
      \If{ {\it conf}(bbox) $> \epsilon$ }
      {
        Seg3d $\gets$ {\sc convert3d}( bbox, D$_{view}$ )\;
        $Cloud_O$ $\gets$ $Cloud_O$ $\cup$ Seg3d\;
      }
    }
    \tcp{point cloud of object O is extracted from depth image according to bbox}
    {\sc Outlier\_Removal}( $Cloud_O$ )\;
    $\trans_O \gets$ {\sc compute\_6dpose}( $Cloud_O$, $M_O$ )\;
    \tcp{6d pose is computed given the point cloud segment and object model}
  }
  \ForEach{ $view \in \trans_{cam}$}
  {
    \{ labels, bboxs \} $\gets$  {\sc project}$( \trans_O, view )$\;
    \tcp{Estimated pose $\trans_O$ is used to generate bounding-boxes in all views}
    dataset $\gets$ dataset $\cup$ (RGB$_{view}$, labels, bboxs)\;
  }
  randObj $\gets$ {\sc sample\_random\_object}( $M_{1:N}$ )\;
  {\sc reconfigure\_object}( randObj )\;
  \tcp{randomly selected objects are moved to pre-specified configuration.}
}
Train Faster-RCNN using the expanded dataset\;
\end{algorithm}

The detector trained using physics-aware simulation is then used to
extract bounding boxes corresponding to each object in the scene (line
7). There might exist a bias in simulation either with respect to
texture or poses, which can lead to imprecise bounding boxes or
complete failure in certain views. For the detection to be considered
for further processing, a threshold is considered on the confidence
value returned by {\tt RCNN} (line 8).

The pixel-wise depth information Seg3d within the confidently detected bounding boxes bbox (line 9) is aggregated in a common point cloud per object $Cloud_O$ given information from multiple views (line 10). The process employs environmental knowledge to clean the aggregated point cloud (line 11). Points outside the resting surface bounds are removed and outlier removal is performed based on k-nearest neighbors and a uniform grid filter.

Several point cloud registration methods were tested for registering the 3D model $M_O$ with the corresponding segmented point cloud $Cloud_O$ (line 12). This included {\tt Super4PCS} (\cite{mellado2014super}), fast global registration (\cite{koltun}) and simply using the principal component analysis ({\tt PCA}) with Iterative Closest Point ({\tt ICP}) (\cite{icp}). The {\sc Super4PCS} algorithm used alongside ICP was found to be the most applicable for the target setup as it is the most robust to outliers and returns a very natural metric for confidence evaluation. {\sc Super4PCS} returns the best rigid alignment according to the Largest Common Pointset ({\tt LCP}). The algorithm searches for the best score, using transformations obtained from four-point congruences. Thus, given enough time, it generates the optimal alignment with respect to the extracted segment.

After the 6-DoF pose is computed for each object, the scene is recreated in the simulator using object models placed at the pose $\trans_O$ and projected to the known camera views (line 14). Bounding boxes are computed on the simulated setup and transferred to the real images. This gives precise bounding box labels for real images in all the views (line 15).

\begin{figure}[ht]
\centering
\includegraphics[width=\linewidth]{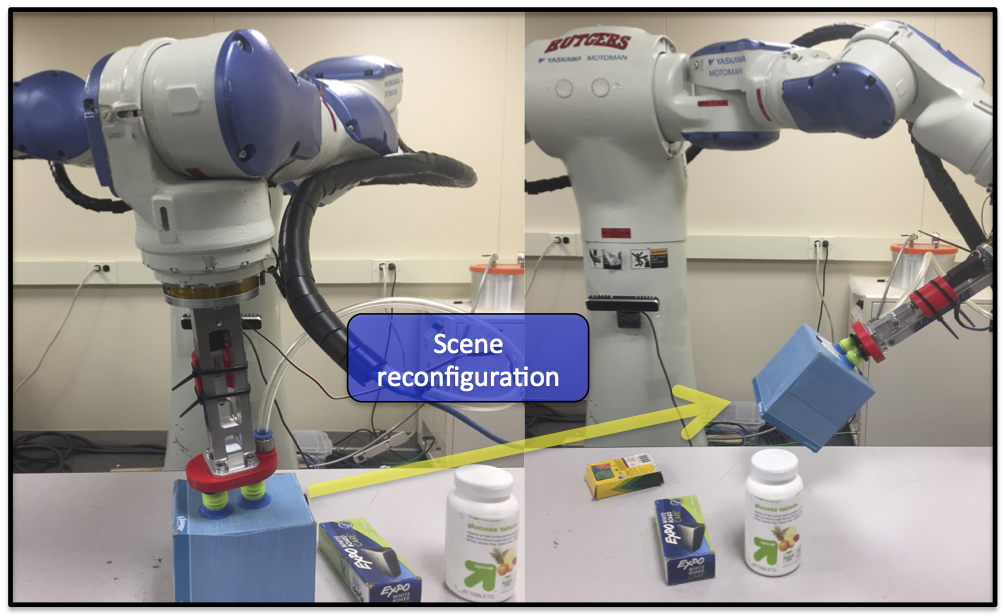}
\vspace{-.3in}
\caption{Manipulator performing scene reconfiguration by moving an object from one configuration on the table to another}
\label{fig:figurelabelx}
\vspace{-.1in}
\end{figure}

To further reduce manual labeling effort, an autonomous scene reconfiguration is performed (lines 16-17). The robot reconfigures the scene with a pick and place manipulation action to iteratively construct new scenes and label them, as in Fig.~\ref{fig:figurelabelx}. For each reconfiguration, the object to be moved is chosen randomly and the final configuration is selected from a set of pre-defined configurations in the workspace.

Finally, the {\tt Faster-RCNN} network is re-trained with the expanded dataset. The factors that prevent this process from a label drift are (1) The network is re-trained with a large number of accurate synthetic data. Thus, the training is immune to some amount of label noise in the self-labeled data. (2) Only the most confident detections from multiple-views are considered and a global search based process for pose estimation is used to obtain the estimates which are eventually used for labeling.

\begin{figure*}[t]
    \centering \includegraphics[width=\textwidth,
      keepaspectratio]{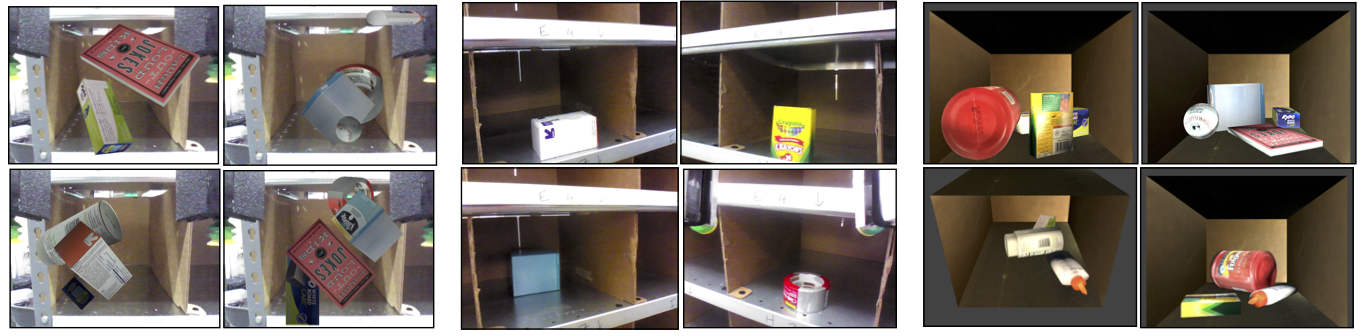}
    \caption{Images from training datasets: (Left) Uniformly sampled
      synthetic data (Center) Training data from the {\tt Shelf\&Tote
        dataset} (\cite{Princeton}) (Right) Dataset generated from the
      proposed physics-aware simulation.}
    \label{fig:trainImages}
\end{figure*}

\section{Evaluation}
\label{sec:evaluation}
This section evaluates several aspects of the proposed approach. It describes the experimental setup, evaluation metrics and compares against baseline alternatives. The evaluations are performed over 4 datasets with different challenges in each, as described in Table~\ref{table:testStats}. The {\tt Shelf\&Tote} dataset (\cite{Princeton}) offers 148 different configurations of objects from the {\tt APC}, placed in bins of a shelf with challenging conditions like occlusions and shiny reflective surfaces of the shelf. In each scene, 2-5 objects are supposed to be detected. {\tt Extended Rutgers RGBD dataset} was created for the purpose of studying the utility of the proposed physics-based pose estimation process. RGB-D images for 42 different configurations of objects were collected and ground truth {\tt 6-DOF} poses were labeled for each object in the image. The dataset contains the same 11 objects from the {\tt APC} as the {\tt Shelf\&Tote} dataset, representing different object geometries. Each scene contains 3 objects to be detected and the object placement is a mix of independent object placements, objects with physical dependencies such as one stacked on/or supporting the other object and occlusions. The dataset was collected using an Intel RealSense sensor mounted over a Motoman robotic manipulator. The {\tt Linemod (LM)} dataset (\cite{hinterstoisser2012model}) is a popular dataset for evaluating pose estimation techniques. For several frames captured from different views, one object instance is labeled per scene. However, the scene has a clutter of other known and unknown objects. \cite{brachmann2014learning} labeled for one such test sequence, the pose of all the known objects (8 object classes) in clutter. This test sequence referred to as the {\tt Linemod-Occluded (LM-O)} dataset has high level of occlusions in several views.

In the first section, the synthetic data generation pipeline and the effect of self-learning will be evaluated on the {\tt Shelf\&Tote} dataset. This will be followed by a detailed summary of performance and accuracy of the pose estimation approach on the {\tt Extended Rutgers RGBD dataset}. Finally, the entire pipeline will be evaluated on the {\tt LM} and {\tt LM-O} dataset according to the recently published benchmark (\cite{hodan2018bop}).

\begin{table}[ht]
\footnotesize\sf\centering
\caption{Statistics for the test datasets. \label{table:testStats}}
\begin{tabular}{ccccc}
\toprule
& Shelf\&Tote & Ext. Rutgers & LM & LM-O\\
\midrule
No. of Objects & 11 & 11 & 8 & 8\\
No. of Scenes & 148 & 42 & 8 & 3\\
No. of Frames & 2220 & 42 & 1600 & 200\\
Objects/scene & 2-5 & 3 & 1 & 8\\
Sensor & \multicolumn{2}{c}{Intel Realsense} &  \multicolumn{2}{c}{Microsoft Kinect}\\
Resolution & \multicolumn{4}{c}{640 X 480}\\
\bottomrule
\end{tabular}
\end{table}

\subsection{Evaluating the dataset generation pipeline}
The dataset generation pipeline is evaluated for the task of bounding-box object detection. A {\tt Faster-RCNN} (\cite{ren2015faster}) based object detector is trained with the datasets generated from different pipelines. The most likely bounding-box prediction for each of the known classes in the scene is considered and a mean average precision ({\tt mAP}) is calculated. The predicted bounding-box is a true positive when the intersection-over-onion ({\tt IoU}) of the predicted bounding box with the ground truth bounding box is greater than a threshold (set to a standard {\tt IoU} value of 0.5).

To study how the object pose distribution affects the training process, different techniques for synthetic data generation are evaluated. The
results of experiments performed on the {\tt Shelf\&Tote} dataset are presented in Table~\ref{table:Synth}.

\begin{table}[h]
\footnotesize\sf\centering
\caption{Evaluating object detection trained with synthetic data. \label{table:Synth}}
\begin{tabular}{lll}
\toprule
Method&mAP (\%)\\
\midrule
\cite{Princeton} (Benchmark)& 75\%\\
Sampled from test data distribution & 69\%\\
Sampled from uniform distribution & 31\%\\
Physics-aware simulation & 64\%\\
Physics-aware simulation + randomized illumination & 70\%\\
\bottomrule
\end{tabular}\\[10pt]
\begin{tabular}{ll}
\toprule
Method&mAP (\%)\\
\midrule
Self-learning (2K images) & 75\% \\
Self-learning (6K images) & 81\% \\
Self-learning (10K images) & 82\% \\
\bottomrule
\end{tabular}
\end{table}

Following is a brief discussion of the dataset generation techniques used for the comparisons:
\subsubsection{Training data generated using test data distribution:}
The objective here is to establish an upper bound for the performance of a detector trained with simulated images. For this purpose, the object detector is trained with the knowledge of pose distribution from the test data. This process consists of estimating the density of the test data with respect to object poses using {\it Kernel Density Estimation}, and generating training data according to this distribution. The sampled scenes were used to train a Faster-RCNN detector, which achieved an accuracy of 69\%.
\subsubsection{Uniformly sampled synthetic data:}
This alternative is a popular technique for generating synthetic data. It uses 3D models of the objects to render their images from several viewpoints sampled on a spherical surface centered at the object. The background image corresponded to the {\tt APC} shelf, on top of which randomly selected objects were pasted at sampled locations. This process allows to simulate occlusions and mask subtraction provides the accurate bounding boxes in these cases. The objects in these images are not guaranteed to have physically realistic poses. This method of synthetic data generation does not perform well on the target task, giving a low accuracy of 31\%.
\subsubsection{Generating training data with physics-aware simulation:}
The accuracy of 64\% achieved by the proposed physics-aware simulator is close to the upper bound. By incorporating the knowledge of the camera pose, resting surface and by using physics simulation, the detector is essentially constraining the distribution of poses to resemble the one from which the test data comes.

The results discussed until now were with respect to a constant lighting condition. As the dataset grows, a dip in the performance is observed. This is expected as the detector overfits with respect to the synthetic texture, which does not mimic real lighting condition. This is not desirable, however. To deal with this issue, the lighting conditions are varied according to the location and color of the light source. This does resolve the problem to some extent but the dataset bias still limits performance to an accuracy of 70\%.

Once a detector is trained with the dataset from simulation, the self-learning pipeline is executed. It is used to automatically label training images from {\tt Shelf\&Tote} dataset. The real images are incrementally added to the simulated dataset to re-train the {\tt Faster-RCNN}. This results in a performance boost of 12\%. This result also outperforms the training process by \cite{Princeton} which uses approximately 15,000 real images labeled using background subtraction. The reason that the proposed method outperforms a large dataset of real training images is that the proposed system can label objects placed in a clutter and not just single instances of objects.

The detector trained from the self-learning pipeline is also evaluated on the task of multi-view pose estimation. Table~\ref{table:poseEstSL}
compares the {\tt Faster-RCNN}-based detector trained with the proposed dataset generation technique to a Fully Convolutional Network
(FCN) trained with dataset generation process from \cite{Princeton}. Different algorithms for estimating the 6d pose is considered and the success is reported by counting the instances when the pose prediction encounters an error in translation less than 5cm and mean error in the rotation less than $15^o$. It is also interesting to note that the success in pose estimation is at par with the success achieved using ground truth bounding boxes.

\begin{table*}[h]
\footnotesize\sf\centering
\caption{Evaluating pose estimation on model learnt from self-learning process.\label{table:poseEstSL}}
\begin{tabular}{lllll}
\toprule
2D-Segmentation Method & 3D-registration Method & Mean-error Rotation (deg) & Mean-error Translation (m) & Success(\%)\\
\midrule
Ground-Truth Bounding-Box & PCA + ICP & 7.65 & 0.02 & 84.8\\
FCN (trained with \cite{Princeton}) & PCA + ICP & 17.3 & 0.06 & 54.6\\
FCN (trained with \cite{Princeton}) & Super4PCS + ICP & 16.8 & 0.06 & 54.2\\
FCN (trained with \cite{Princeton}) & fast-global-registration & 18.9 & 0.07 & 43.7\\
Faster-RCNN (Proposed training) & PCA + ICP & {\bf 8.50} & 0.03 & {\bf 79.4}\\
Faster-RCNN (Proposed training) & Super4PCS + ICP & 8.89 & {\bf 0.02} & 75.0\\
Faster-RCNN (Proposed training) & fast-global-registration & 14.4 & 0.03 & 58.9\\
\bottomrule
\end{tabular}\\[10pt]
\end{table*}

\begin{figure}[t]
    \centering \includegraphics[width=\linewidth, keepaspectratio]{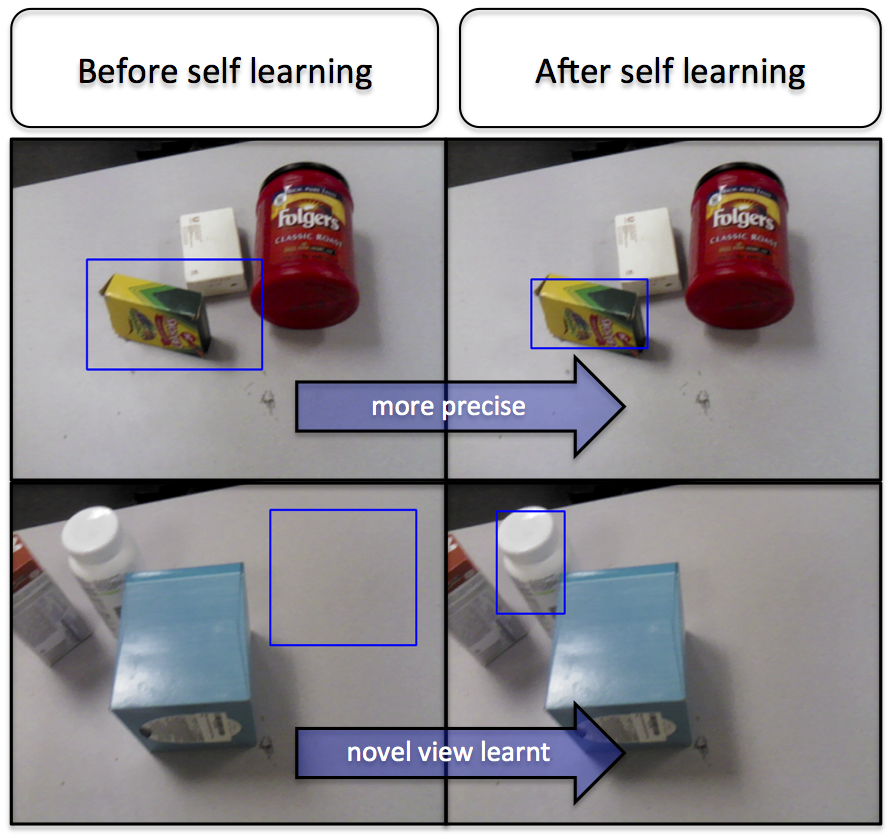}
    \caption{Results of object detection before and after training with the self-learning process. The detector learns to predict more precise bounding boxes. It can also detect objects better from novel views.}
    \label{fig:resultsSL}
\end{figure}

\subsection{Evaluating the Search-based Pose Estimation}

\begin{figure}[t]
    \centering \includegraphics[width=0.9\linewidth,
      keepaspectratio]{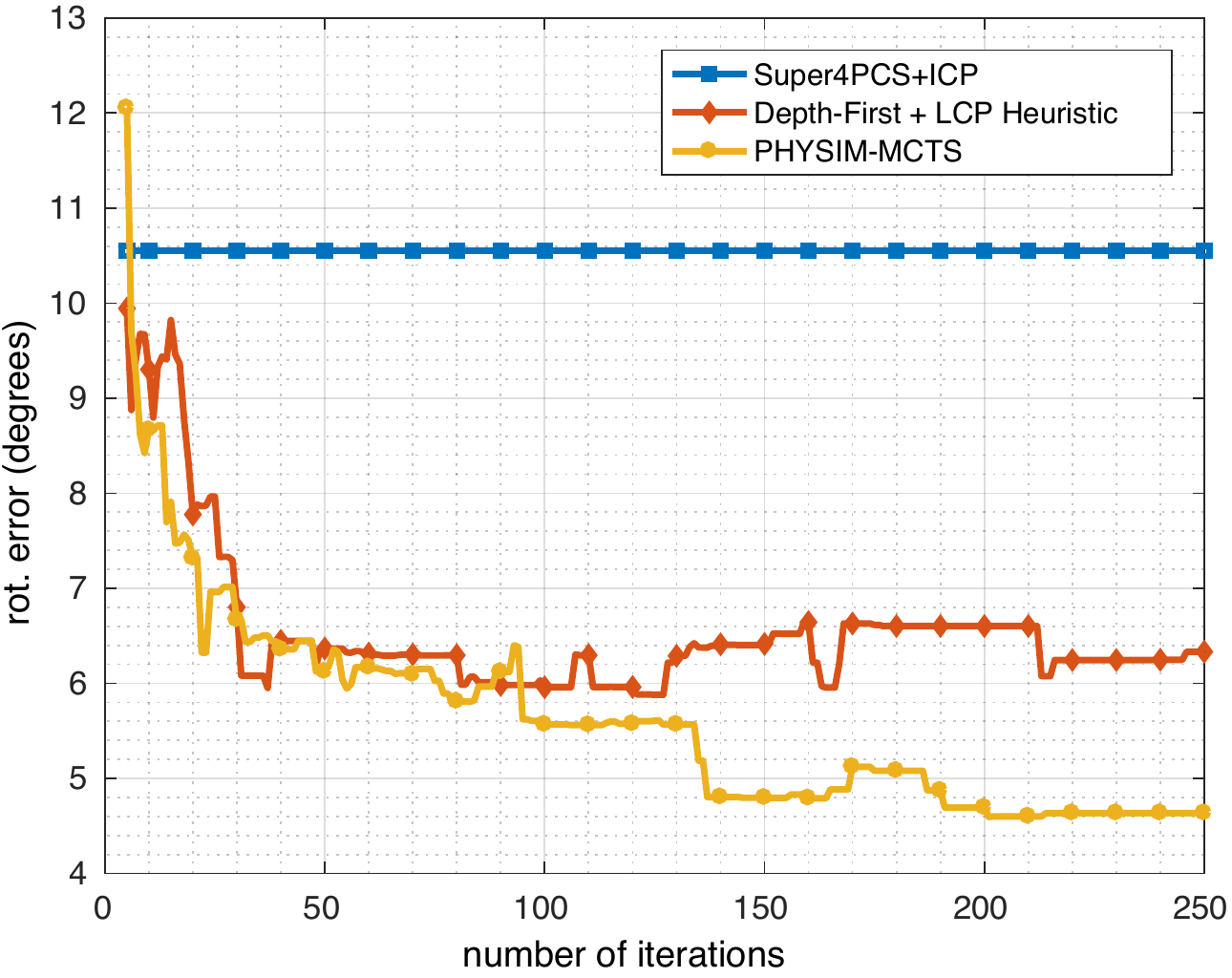}
    \caption{Rotation error in degrees as a function of the number of
      iterations.}

    \label{fig:rotErr}
\end{figure}

\begin{figure}[h!]    
    \centering \includegraphics[width=0.9\linewidth,
      keepaspectratio]{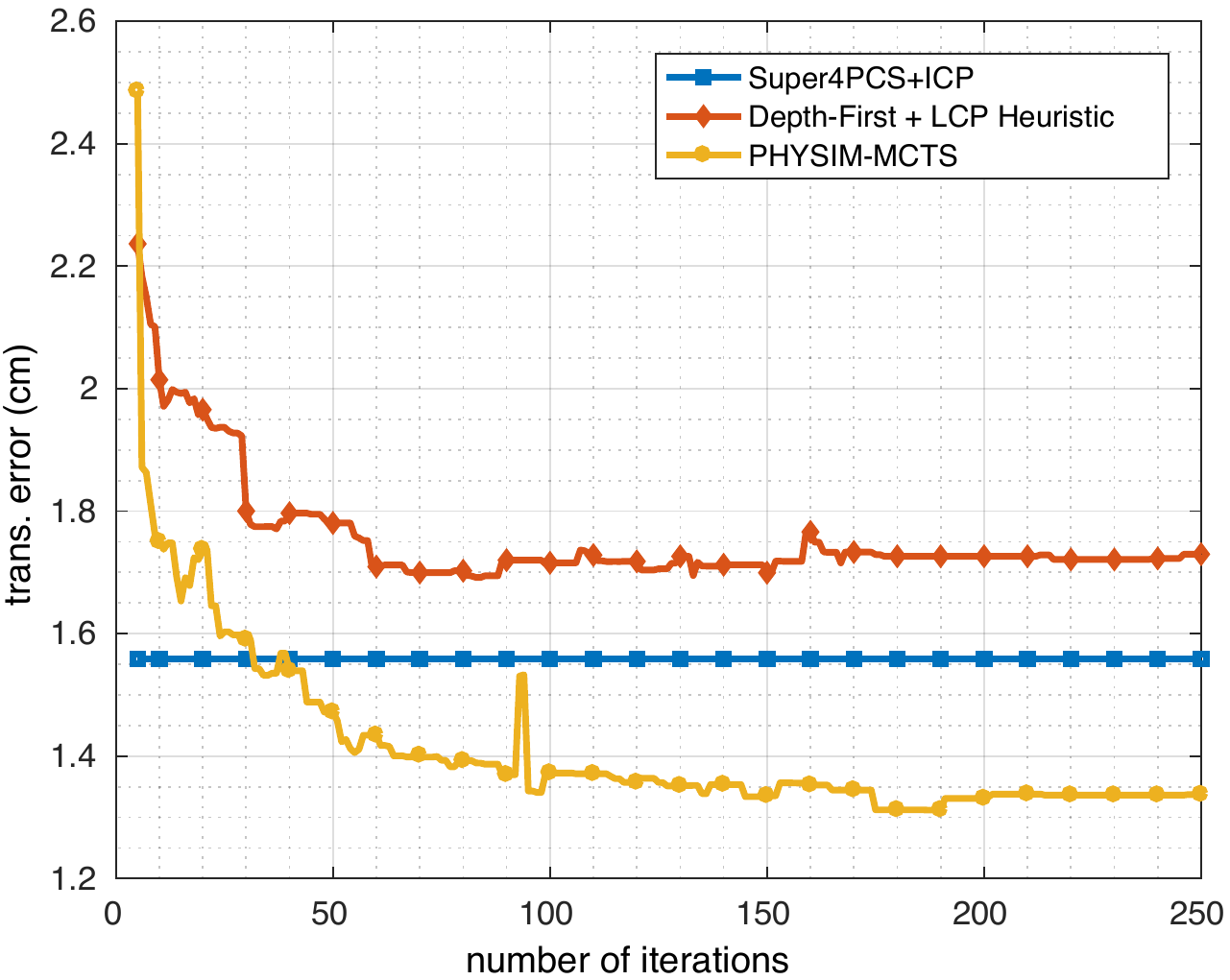}    
    \caption{Translation error in cm as a function of the number of iterations.}
    \label{fig:transErr}
\end{figure}

In this section, the proposed Monte Carlo tree search (MCTS) based algorithm is evaluated over the {\tt Extended Rutgers RGBD dataset}. The scenes in the dataset express three different levels of interaction between objects, namely, independent object placement where an object is physically independent of the rest of objects, two-object dependencies where an object depends on another, and three object dependencies where an object depends on two other objects.

The evaluation is performed by computing the error in translation, which is the Euclidean distance of an object's center compared to its ground truth center (in centimeters). The error in rotation is computed by first transforming the computed rotation to the frame attached to the object at ground truth. The rotation error is the average of the roll, pitch and yaw angles (in degrees) of the transformation between the returned rotation and the ground truth one, while taking into account the object's symmetries, which may allow multiple correct answers. The results provide the mean of the errors of all the objects in the dataset.

\begin{table*}[ht]
\small\sf\centering
\caption{Comparing our approach with different pose estimation techniques.\label{table:poseEst}}
\resizebox{\textwidth}{!}{\begin{tabular}{c |c c|c c|c c|c c}
\toprule
\centering{\bf Method} & \multicolumn{2}{c|}{No Dependencies} & \multicolumn{2}{c|}{2-objects Dependencies} & \multicolumn{2}{c|}{3-objects Dependencies} & \multicolumn{2}{c}{All}\\
\midrule
& Rot. Err. & Trans. Err. & Rot. Err. & Trans. Err. & Rot. Err. & Trans. Err.&Rot. Err. & Trans. Err. \\
APC-Vision-Toolbox & 15.5$^{\circ}$ & 3.4 cm & 26.3$^{\circ}$ & 5.5 cm& 17.5$^{\circ}$ & 5.0 cm& 21.2$^{\circ}$ & 4.8 cm\\
faster-RCNN + Super4PCS + ICP & 2.4$^{\circ}$ & 0.8 cm& 14.8$^{\circ}$ & 1.7 cm& 12.1$^{\circ}$ & 2.1 cm& 10.5$^{\circ}$ & 1.5 cm\\
PHYSIM-Heuristic (depth + LCP) & 2.8$^{\circ}$ & 1.1 cm& 5.8$^{\circ}$ & 1.4 cm& 12.5$^{\circ}$ & 3.1 cm& 6.3$^{\circ}$ & 1.7 cm\\
PHYSIM-MCTS (proposed approach) & 2.3$^{\circ}$ & 1.1 cm& 5.8$^{\circ}$ & 1.2 cm& 5.0$^{\circ}$ & 1.8 cm& {\bf 4.6$^{\circ}$} & {\bf 1.3 cm}\\
\bottomrule
\end{tabular}}\\[10pt]
\end{table*}

\begin{figure*}[ht]
    \centering \includegraphics[width=0.88\textwidth, keepaspectratio]{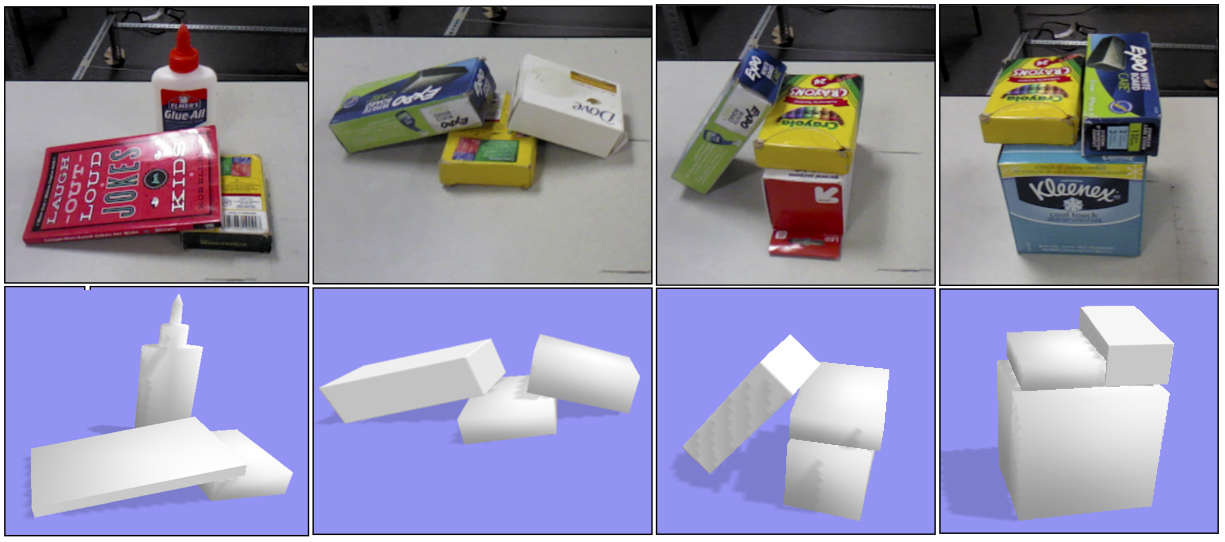}
    \caption{Example images from {\tt Extended Rutgers RGBD dataset}
      and accompanying results from the Monte Carlo Tree Search
      process. The results are visualized in the light-weight physics
      engine ({\tt Bullet}) which plays an integral part in performing
      the local optimization in this pipeline and ensures that the
      returned results are physically stable configurations.}
    \label{fig:mctsRes}
\end{figure*}

The evaluation was first performed for methods that reason about one object at a time, i.e. methods that do not perform any scene-level reasoning. These approaches trust the segments returned by the object segmentation module and perform model matching followed by local refinement to compute object poses. The results of performing pose estimation over the collected dataset with some of these techniques are presented in Table~\ref{table:poseEst}. \cite{Princeton} developed a system for pose estimation towards the Amazon Picking Challenge 2016. The system uses a Fully Convolutional Network ({\tt FCN}) to get pixel level segmentation of objects in the scene, then uses Principal Component Analysis ({\tt PCA}) for pose initialization, followed by {\tt ICP} to get the final object pose. This system was designed for shelf and tote environments and often relies on multiple views of the scene. Thus, the high error in pose estimates could be attributed to the low recall percentage in retrieving object segment achieved by the semantic segmentation method, which in turn resulted in the segment not having enough information to compute a unique pose estimate. The second system tested uses a {\tt Faster-RCNN}-based object detector trained with the {\tt physim-dataset-generator} as described before. The point cloud segments extracted from the bounding box detections were used to perform pose estimation using two different approaches: i) {\tt PCA} followed by {\tt ICP} and ii) {\tt Super4PCS} followed by {\tt ICP} \cite{icp}. Even though the detector succeeded in providing a high recall object segment on most occasions, in the best case the mean rotation error using local approaches was still high ({10.5}$^{\circ}$). This was sometimes due to bounding boxes containing parts of other object segments, or due to occlusions. Reasoning only at a local object-level does not resolve these issues.

The proposed search framework was used to perform pose estimation on
the dataset. In each scene, the dependency graph structure was used to
get the order of object placement and initialize the independent
search trees. Then, the object detection was performed using {\tt
  Faster-RCNN} and congruent set matching was used to generate pose
candidates, which were clustered to get 25 representatives per
object. The search is performed over the combined set of object
candidates and the output of the search is an anytime pose estimate
based on the best rendering score. The stopping criterion for the
searches was defined by a maximum number of node expansions in the
tree, set to 250, where each expansion corresponds to a physics
simulation with {\tt Bullet} and a rendering with {\tt OpenGL}, with a
mean expansion time of $\sim0.2$ secs per node. The search was
initially performed using a depth-first heuristic combined with the
LCP score returned by the Super4PCS for the pose candidates. The
results from this approach, {\tt PHYSIM-Heuristic (depth + LCP)}, are
shown in Table~\ref{table:poseEst}, which indicates that it might be
useful to use these heuristics if the tree depth is low (one and two
object dependencies). As the number of object dependencies grows,
however, one needs to perform more exploration. For three-object
dependencies, when using 250 expansions, this heuristic search
provided poor performance. The UCT Monte Carlo Tree Search was used to
perform the search, with upper confidence bounds to trade off
exploration and exploitation. The exploration parameter was set to a
high value ($\alpha = 5000$), to allow the search to initially look
into more branches while still preferring the ones that give a high
rendering score. This helped in speeding up the search process
significantly, and a much better solution could be reached within the
same time. The plots in Fig.~\ref{fig:rotErr} and
Fig.~\ref{fig:transErr} captures the anytime results from the two
heuristic search approaches. Fig.~\ref{fig:mctsRes} shows some of the
images from the {\tt Extended Rutgers RGBD dataset} and the
corresponding results from the UCT Monte Carlo Tree Search process.
\begin{figure*}[ht]
    \centering \includegraphics[width=\linewidth, keepaspectratio]{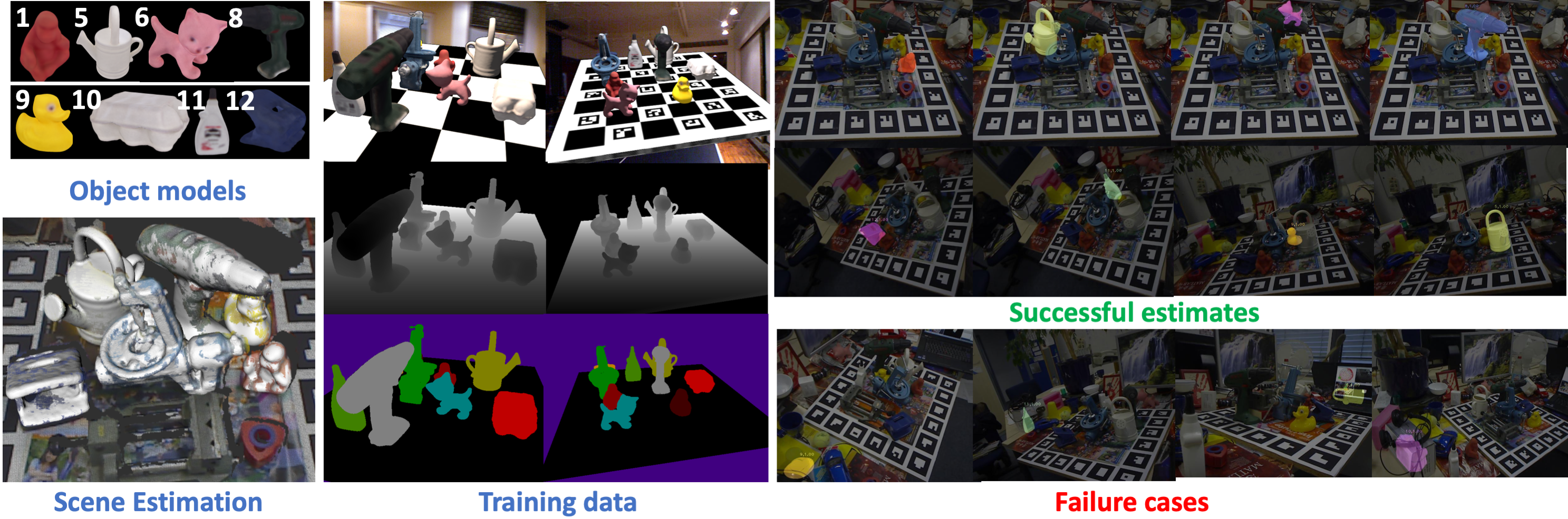}
    \caption{Performing pose estimation over {\tt Linemod} and {\tt Linemod-Occluded} dataset. The visualizations demonstrate (Left) the object models and final result of the pose estimation process on RGB-D data. (Middle) the training data generated from the proposed pipeline and (Right) some instances of successfull estimates as well as failure cases on {\tt Linemod-Occluded} dataset with high level of occlusion.}
    \label{fig:linemodTest}
\end{figure*}
\begin{table*}[ht]
\ssmall\centering
\begin{center}
\begin{tabular}{lccccccccc}
\hline
& \multicolumn{9}{c}{LINEMOD (recall \%)}\\
\hline
& 1 & 5 & 6 & 8 & 9 & 10 & 11 & 12 & All\\
\hline
\cite{hodavn2015detection} & 91 & 91 & 97 & 69 & 90 & 97 & 81 & 79 & 86.9\\
\rowcolor{Gray}
MCTS & 93 & 90 & 87 & 90 & 80 & 97 & 80 & 65 & 85.3\\
\cite{vidal20186d} & 89 & 92 & 96 & 89 & 87 & 97 & 59 & 69 & 84.8\\
\cite{drost2010model} & 86 & 93 & 87 & 92 & 66 & 96 & 53 & 67 & 80.0\\
\cite{drost2010model} (Edge) & 77 & 98 & 94 & 96 & 45 & 94 & 68 & 66 & 79.8\\
\cite{brachmann2016uncertainty} & 91 & 86 & 90 & 72 & 85 & 79 & 46 & 67 & 77.0\\
\cite{hodavn2015detection} (NR) & 91 & 66 & 87 & 49 & 92 & 90 & 65 & 63 & 75.4\\
\cite{brachmann2014learning} & 74 & 88 & 66 & 81 & 69 & 66 & 50 & 75 & 71.1\\
\cite{Kehl:2016aa} & 60 & 79 & 68 & 68 & 42 & 91 & 45 & 42 & 61.9\\
\cite{buch2017rotational} (ppfh) & 77 & 84 & 60 & 59 & 75 & 67 & 24 & 39 & 60.6\\
\cite{buch2017rotational} (si) & 40 & 81 & 47 & 8 & 36 & 43 & 18 & 3 & 34.5\\
\cite{buch2017rotational} (ecsad) & 31 & 66 & 3 & 0 & 9 & 49 & 1 & 0 & 19.9\\
\cite{tejani2014latent} & 36 & 1 & 0 & 11 & 1 & 70 & 27 & 0 & 18.3\\
\cite{buch2016local} (ppfh) & 11 & 3 & 7 & 7 & 18 & 12 & 4 & 3 & 8.1\\
\cite{buch2017rotational} (shot) & 3 & 9 & 4 & 3 & 2 & 10 & 1 & 0 & 4.0\\
\cite{buch2016local} (ecsad) & 2 & 5 & 0 & 4 & 5 & 8 & 0 & 0 & 3.0\\
SL-MCTS & - & - & - & - & - & - & - & - & -\\
\end{tabular}%
\begin{tabular}{lccccccccc}
\hline
& \multicolumn{9}{c}{LINEMOD-Occluded (recall \%)}\\
\hline
& 1 & 5 & 6 & 8 & 9 & 10 & 11 & 12 & All\\
\hline
\rowcolor{Gray}
SL-MCTS & 50 & 71 & 43 & 68 & 72 & 46 & 33 & 66 & 60.3\\
\cite{vidal20186d} & 66 & 81 & 46 & 65 & 73 & 43 & 26 & 64 & 59.3\\
\rowcolor{Gray}
MCTS & 48 & 59 & 35 & 78 & 71 & 48 & 32 & 65 & 58.4\\
\cite{drost2010model} & 62 & 75 & 39 & 70 & 57 & 46 & 26 & 57 & 55.4\\
\cite{drost2010model} (Edge) & 47 & 82 & 46 & 75 & 42 & 44 & 36 & 57 & 55.0\\
\cite{brachmann2016uncertainty} & 64 & 65 & 44 & 68 & 71 & 3 & 32 & 61 & 52.0\\
\cite{hodavn2015detection} & 54 & 66 & 40 & 26 & 73 & 37 & 44 & 68 & 51.4\\
\cite{brachmann2014learning} & 50 & 48 & 27 & 44 & 60 & 6 & 30 & 62 & 41.5\\
\cite{buch2017rotational} (ppfh) & 59 & 63 & 18 & 35 & 60 & 17 & 5 & 30 & 37.0\\
\cite{hodavn2015detection} (NR) & 47 & 35 & 24 & 12 & 63 & 9 & 32 & 53 & 34.4\\
\cite{Kehl:2016aa} & 39 & 47 & 24 & 30 & 48 & 14 & 13 & 49 & 33.9\\
\cite{buch2017rotational} (si) & 54 & 63 & 11 & 2 & 16 & 9 & 1 & 3 & 20.4\\
\cite{buch2017rotational} (ecsad) & 29 & 29 & 0 & 0 & 7 & 8 & 1 & 0 & 9.6\\
\cite{tejani2014latent} & 26 & 2 & 0 & 1 & 0 & 0 & 10 & 0 & 4.5\\
\cite{buch2016local} (ppfh) & 4 & 0 & 0 & 2 & 11 & 1 & 1 & 1 & 2.3\\
\cite{buch2017rotational} (shot) & 2 & 7 & 0 & 0 & 1 & 1 & 1 & 0 & 1.5\\
\cite{buch2016local} (ecsad) & 1 & 3 & 0 & 2 & 2 & 0 & 0 & 0 & 1.0\\
\end{tabular}
\end{center}
\caption{Evaluating the performance of the proposed search process on the {\tt LINEMOD} and {\tt LINEMOD-Occluded} dataset according to the recent benchmark (\cite{hodan2018bop}) in pose estimation.}
\label{table:linemod}
\end{table*}
To study the effect of training on the pose estimation process, an
experiment was performed which utilizes the ground-truth segmentation
of objects and performs the {\tt PHYSIM-MCTS} to generate object
poses. This resulted in a rotation error of 2.94$^{\circ}$ and a
translation error of 0.7cm. This is not significantly different from
the results with the proposed process which indicates that the
bounding-box detector trained from the autonomous dataset generation
pipeline already provides enough information for this pose estimation
process. Some of the reasons for failures in pose estimation when
evaluated with ground truth segmentation were found to be because of
pose averaging when using cluster centers as pose representatives
resulting in the failure of local optimization and when the depth sensor did
not return points for some reflective surfaces.

\subsection{Evaluating over Benchmark for Pose Estimation}

In this section, the entire pipeline is evaluated over the {\tt Linemod} (\cite{hinterstoisser2012model}) and the {\tt Linemod-Occluded} (\cite{brachmann2014learning}) datasets. Evaluation is performed according to the benchmark (\cite{hodan2018bop}) for 8 objects as shown in Fig.~\ref{fig:linemodTest} which have corresponding ground-truth pose labels in both the datasets. The accuracy is measured in terms of the Visual Surface Discrepancy (VSD) metric as defined in \cite{hodan2018bop} with a misalignment tolerance of $\tau=20mm$ and correctness threshold $\theta=0.3$. Given these parameters, the error is calculated by rendering the object model at the predicted and the ground-truth pose as depth maps S and S'. These are compared to the actual depth map of the image to obtain visibility masks V and V' and the error is calculated as, 
\begin{align*}
e_{vsd} = \underset{p \in V \cap V'}{avg}
\begin{cases}
0 \text{ if } p \in V \cap V' \land \mid S(p) - S'(p) \mid < \tau \cr
1, \text{otherwise.}
\end{cases}
\end{align*}

A pose is counted as correct if $e_{vsd} < \theta$. Finally the recall rate per object and over the entire dataset are presented in the Table~\ref{table:linemod}. 

To compare the proposed approach, first a synthetic training data was generated based on the developed pipeline. Some examples of the generated images are shown in Fig.~\ref{fig:linemodTest}. Overall 30,000 RGB and corresponding depth images are generated along with per pixel class labels. To generate this dataset, the intrinsic camera parameters, object texture and pose of the table are kept constant. The pose of the object over the table is varied randomly over x,y position and yaw while the rest of the pose parameters are kept constant. Finally, physics simulation is applied to get a physically-consistent scene which is rendered from 20 different viewpoints. The viewpoints are sampled randomly from a hemisphere of radius varying in a range similar to the test dataset. The camera sampling policy and range values are similar to the one used for generating the training data in \cite{hodan2018bop}. Other scene parameters like light position, light color, object material emission, background and texture of the table are varied randomly within a pre-specified domain. 

A Fully-convolutional network (FCN) (\cite{shelhamer2016fully}) is trained with the generated data to obtain pixel-level classification and the output is used to guide the pose estimation process. The choice of using an {\tt FCN} instead of {\tt Faster-RCNN} was due to the fact that several unknown objects are present in the scene and predicting one definite location for an object in the scene would reduce the recall rate for the recognition task. In the {\tt Linemod} dataset only one object needs to be estimated in each frame. To perform this task, first the pose of the table is computed using a RANSAC-based process and the direction of gravity is assumed to be perpendicular to the surface of the table. Then, 50 pose candidates are considered for the object based on the segmentation output and each of these are locally-optimized based on physics simulation and ICP in the {\tt MCTS} process. Finally, the score is computed to select the best candidate. Due to the presence of unmodeled clutter, the optimization cost cannot assume that the entire scene can be explained by the estimated pose of known objects. Thus, the optimization cost for this dataset is set so as to maximize the alignment of the rendered depth map of the object at the predicted pose with the observed depth map. The alignment is computed with a distance threshold of 10mm and a surface normal tolerance of 30 degrees. The surface normal is used to avoid cases where the objects are falsely assigned to parts of large flat surfaces.

On the {\tt Linemod-Occluded} dataset pose for 8 objects need to be estimated in every image with high level of occlusion. Two separate tests are performed on this dataset. In the first experiment the {\tt FCN} is trained with just synthetic data from the proposed pipeline and the output is used to guide the {\tt MCTS} process to estimate the pose for all 8 objects present in the scene. An example of the prediction is visualized in at the bottom-left of Fig.~\ref{fig:linemodTest}. In the second experiment, the {\tt FCN} is re-trained with additional images from the {\tt Linemod} dataset which are labeled using the confident estimates from the pose estimation over the entire dataset and projected to all the different views. Note that in this case only the segment corresponding to one object could be extracted from each image of the {\tt Linemod} dataset, so a mask is used during the training process to only use that small part of the image which corresponds to the object and ignores the rest. This presents only positive samples for training on real-data and thus not a very significant improvement can be seen from this task. The performance corresponding to this experiment is referred to as the {\tt MCTS-SL} in Table~\ref{table:linemod}.

Overall, the proposed approach achieves state-of-the-art performance on both of these datasets. On the {\tt Linemod} dataset, the proposed pipeline which is just trained on synthetic data achieves 85.3\% accuracy that is just slightly below the template matching work of \cite{hodavn2015detection} (86.9\%) in terms of overall success. Although template matching works well in cases of less occlusion, it fails to achieve a high recall on occluded datasets. Thus, on the {\tt Linemod-Occluded} dataset, our proposed approach achieves the highest recall rate of 60.3\% when the entire pipeline is used. When the self-learning component is not used, the performance is still just slightly below the top performing method of \cite{vidal20186d}. 

Some examples of the successful estimates and failure conditions on this dataset are presented in Fig.~\ref{fig:linemodTest}. One of the cases for failure is the presence of unmodeled objects on which the target object are physically dependent (Eggbox object in the bottom-right corner of the figure). The other failure case is that of object models getting good alignment scores with similar looking and large surfaces in the image (first three failure cases in the figure).

\subsection{Limitations}

One of the limitations of global reasoning, as is in this approach, is the time required for computing and searching over an extensive hypotheses set. Particularly, due to the hierarchical clustering approach that was adapted to consider object specific distances, the hypotheses generation time for an object can be in the order of multiple seconds. The search process, which seemed to converge to good solutions with 150 expansions for three-object dependencies, takes approximately 30 seconds. Nevertheless, both of these processes are highly parallelizable. Future work can perform the hypotheses generation and the search with parallel computing. Another limitation of this work in the current form is the assumption that the objects are non-transparent and rigid. For transparent objects, this is due to the lack of depth data on the surface of these objects.

\section{Discussion}
\label{sec:conclusion}
This work provides a comprehensive framework for 6 \dof\ pose
estimation of objects placed in clutter. It leverages the advantages
of recent success in deep learning without the need for any manual
effort in data collection and labeling. It offers a novel way of
performing pose estimation for objects placed in clutter by
efficiently searching for the best scene explanation over the space of
physically consistent scene configurations. It also provides a method
to construct these sets of scene configurations by using
state-of-the-art object detection and model registration techniques,
which by themselves are not sufficient to give a desirable pose
estimate for objects. The evaluations indicate significant performance
improvement in both the tasks of object detection and pose estimation
using the proposed approach. The limitations mentioned in the previous
section encourage future work on fast and robust hypotheses generation
and developing a method to systematically and quickly cluster object
poses in SE(3), while taking into consideration the symmetries of
objects. There is also a wide interest in bridging the domain gap between simulated 
and real images by domain randomization (\cite{tobin2017domain}) or with a generative 
learning technique (\cite{shrivastava2017learning}). The current work could 
leverage such techniques to provide an even better initialization to this process.

\bibliographystyle{format/SageH}
\bibliography{bib/physics_perception.bbl}

\begin{thebibliography}{86}
\providecommand{\natexlab}[1]{#1}
\providecommand{\url}[1]{\texttt{#1}}
\providecommand{\urlprefix}{URL }
\expandafter\ifx\csname urlstyle\endcsname\relax
  \providecommand{\doi}[1]{DOI:\discretionary{}{}{}#1}\else
  \providecommand{\doi}{DOI:\discretionary{}{}{}\begingroup
  \urlstyle{rm}\Url}\fi

\bibitem[{Aiger et~al.(2008)Aiger, Mitra and Cohen-Or}]{aiger20084}
Aiger D, Mitra NJ and Cohen-Or D (2008) {4-points Congruent Sets for Robust
  Pairwise Surface Registration}.
\newblock In: \emph{ACM Transactions on Graphics (TOG)}, volume~27. ACM, p.~85.

\bibitem[{Akizuki and Hashimoto(2016)}]{akizuki2016physical}
Akizuki S and Hashimoto M (2016) Physical reasoning for 3d object recognition
  using global hypothesis verification.
\newblock In: \emph{Computer Vision--ECCV 2016 Workshops}. Springer, pp.
  595--605.

\bibitem[{Aldoma et~al.(2012{\natexlab{a}})Aldoma, Marton, Tombari, Wohlkinger,
  Potthast, Zeisl, Rusu, Gedikli and Vincze}]{aldoma2012tutorial}
Aldoma A, Marton ZC, Tombari F, Wohlkinger W, Potthast C, Zeisl B, Rusu RB,
  Gedikli S and Vincze M (2012{\natexlab{a}}) Tutorial: Point cloud library:
  Three-dimensional object recognition and 6 dof pose estimation.
\newblock \emph{IEEE Robotics \& Automation Magazine} 19(3): 80--91.

\bibitem[{Aldoma et~al.(2012{\natexlab{b}})Aldoma, Tombari, Di~Stefano and
  Vincze}]{aldoma2012global}
Aldoma A, Tombari F, Di~Stefano L and Vincze M (2012{\natexlab{b}}) A global
  hypotheses verification method for 3d object recognition.
\newblock In: \emph{European Conference on Computer Vision}. Springer.

\bibitem[{Aldoma et~al.(2013)Aldoma, Tombari, Prankl, Richtsfeld, Di~Stefano
  and Vincze}]{aldoma2013multimodal}
Aldoma A, Tombari F, Prankl J, Richtsfeld A, Di~Stefano L and Vincze M (2013)
  Multimodal cue integration through hypotheses verification for rgb-d object
  recognition and 6dof pose estimation.
\newblock In: \emph{Robotics and Automation (ICRA), 2013 IEEE International
  Conference on}. IEEE, pp. 2104--2111.

\bibitem[{Arthur and Vassilvitskii(2007)}]{arthur2007k}
Arthur D and Vassilvitskii S (2007) k-means++: The advantages of careful
  seeding.
\newblock In: \emph{Proceedings of the eighteenth annual ACM-SIAM symposium on
  Discrete algorithms}. Society for Industrial and Applied Mathematics, pp.
  1027--1035.

\bibitem[{Ballard(1981)}]{ballard1981generalizing}
Ballard DH (1981) Generalizing the hough transform to detect arbitrary shapes.
\newblock \emph{Pattern recognition} 13(2): 111--122.

\bibitem[{Besl and McKay(1992)}]{icp}
Besl PJ and McKay ND (1992) {Method for Registration of 3D Shapes}.
\newblock \emph{International Society for Optics and Photonics} .

\bibitem[{Birdal and Ilic(2015)}]{birdal2015point}
Birdal T and Ilic S (2015) Point pair features based object detection and pose
  estimation revisited.
\newblock In: \emph{3D Vision (3DV), 2015 International Conference on}. IEEE,
  pp. 527--535.

\bibitem[{Bo et~al.(2014)Bo, Ren and Fox}]{Bo:2014aa}
Bo L, Ren X and Fox D (2014) {Learning hierarchical sparse features for RGB-(D)
  object recognition}.
\newblock \emph{International Journal of Robotics Research (IJRR)} 33(4).

\bibitem[{Bouazix et~al.(2013)Bouazix, Tagliasacchi and Pauly}]{Bouazix:2013aa}
Bouazix S, Tagliasacchi A and Pauly M (2013) {Sparse Iterative Closest Point}.
\newblock \emph{{Computer Graphics Forum (Symposium on Geometry Processing)}}
  32(5): 1--11.

\bibitem[{Brachmann et~al.(2014)Brachmann, Krull, Michel, Gumhold, Shotton and
  Rother}]{brachmann2014learning}
Brachmann E, Krull A, Michel F, Gumhold S, Shotton J and Rother C (2014)
  Learning 6d object pose estimation using 3d object coordinates.
\newblock In: \emph{European conference on computer vision}. Springer, pp.
  536--551.

\bibitem[{Brachmann et~al.(2016)Brachmann, Michel, Krull, Ying~Yang, Gumhold
  et~al.}]{brachmann2016uncertainty}
Brachmann E, Michel F, Krull A, Ying~Yang M, Gumhold S et~al. (2016)
  Uncertainty-driven 6d pose estimation of objects and scenes from a single rgb
  image.
\newblock In: \emph{Proceedings of the IEEE Conference on Computer Vision and
  Pattern Recognition}. pp. 3364--3372.

\bibitem[{Buch et~al.(2017)Buch, Kiforenko and Kraft}]{buch2017rotational}
Buch AG, Kiforenko L and Kraft D (2017) Rotational subgroup voting and pose
  clustering for robust 3d object recognition.
\newblock In: \emph{2017 IEEE International Conference on Computer Vision
  (ICCV)}. IEEE, pp. 4137--4145.

\bibitem[{Buch et~al.(2016)Buch, Petersen and Kr{\"u}ger}]{buch2016local}
Buch AG, Petersen HG and Kr{\"u}ger N (2016) Local shape feature fusion for
  improved matching, pose estimation and 3d object recognition.
\newblock \emph{SpringerPlus} 5(1): 297.

\bibitem[{Cao et~al.(2016)Cao, Sheikh and Banerjee}]{Cao:2016aa}
Cao Z, Sheikh Y and Banerjee NK (2016) {Real-time Scalable 6DOF Pose Estimation
  for Textureless Objects}.
\newblock In: \emph{{IEEE} International Conference on Robotics and Automation
  (ICRA)}. pp. 2441--2448.

\bibitem[{Cheng et~al.(2013)Cheng, Chen, Martin, Lai and Wang}]{Cheng:2013aa}
Cheng ZQ, Chen Y, Martin R, Lai YK and Wang A (2013) {Supermatching: Feature
  Matching using Supersymmetric Geometric Constraints}.
\newblock In: \emph{IEEE TVCG}, volume~19. p.~11.

\bibitem[{Chetverikov et~al.(2002)Chetverikov, Svirko, Stepanov and
  Krsek}]{chetverikov2002trimmed}
Chetverikov D, Svirko D, Stepanov D and Krsek P (2002) The trimmed iterative
  closest point algorithm.
\newblock In: \emph{Pattern Recognition, 2002. Proceedings. 16th International
  Conference on}, volume~3. IEEE, pp. 545--548.

\bibitem[{Choi and Christensen(2012)}]{choi20123d}
Choi C and Christensen HI (2012) 3d pose estimation of daily objects using an
  rgb-d camera.
\newblock In: \emph{Intelligent Robots and Systems (IROS), 2012 IEEE/RSJ
  International Conference on}. IEEE, pp. 3342--3349.

\bibitem[{Collet et~al.(2011)Collet, Martinez and Srinivasa}]{Collet:2011aa}
Collet A, Martinez M and Srinivasa S (2011) {The MOPED framework: Object
  Recognition and Pose Estimation for Manipulation}.
\newblock \emph{International Journal of Robotics Research (IJRR)} 30(10):
  1284--1306.

\bibitem[{Correll et~al.(2016)Correll, Bekris, Berenson, Brock, Causo, Hauser,
  Osada, Rodriguez, Romano and Wurman}]{Correll:2016aa}
Correll N, Bekris KE, Berenson D, Brock O, Causo A, Hauser K, Osada K,
  Rodriguez A, Romano J and Wurman P (2016) {Analysis and Observations From the
  First Amazon Picking Challenge}.
\newblock \emph{IEEE Trans. on Automation Science and Engineering (T-ASE)} .

\bibitem[{Dhillon et~al.(2004)Dhillon, Guan and Kulis}]{dhillon2004kernel}
Dhillon IS, Guan Y and Kulis B (2004) Kernel k-means: spectral clustering and
  normalized cuts.
\newblock In: \emph{Proceedings of the tenth ACM SIGKDD international
  conference on Knowledge discovery and data mining}. ACM, pp. 551--556.

\bibitem[{Drost and Ilic(2012)}]{drost20123d}
Drost B and Ilic S (2012) {3D Object Detection and Localization using
  Multimodal Point Pair Features}.
\newblock In: \emph{Second International Conference on 3D Imaging, Modeling,
  Processing, Visualization and Transmission (3DIMPVT)}. pp. 9--16.

\bibitem[{Drost et~al.(2010)Drost, Ulrich, Navab and Ilic}]{drost2010model}
Drost B, Ulrich M, Navab N and Ilic S (2010) {Model Globally, Match Locally:
  Efficient and Robust 3D Object Recognition}.
\newblock In: \emph{IEEE Conference on Computer Vision and Pattern Recognition
  (CVPR)}. pp. 998--1005.

\bibitem[{Durrant-Whyte and Bailey(2006)}]{durrant2006simultaneous}
Durrant-Whyte H and Bailey T (2006) Simultaneous localization and mapping: part
  i.
\newblock \emph{IEEE robotics \& automation magazine} 13(2): 99--110.

\bibitem[{Erkent et~al.(2016)Erkent, Shukla and Piater}]{Erkent:2016aa}
Erkent O, Shukla D and Piater J (2016) {Integration of Probabilistic Pose
  Estimates from Multiple Views}.
\newblock In: \emph{European Conference on Computer Vision (ECCV)}.

\bibitem[{Fischler and Bolles(1981{\natexlab{a}})}]{fischler1981random}
Fischler MA and Bolles RC (1981{\natexlab{a}}) Random sample consensus: a
  paradigm for model fitting with applications to image analysis and automated
  cartography.
\newblock \emph{Communications of the ACM} 24(6): 381--395.

\bibitem[{Fischler and Bolles(1981{\natexlab{b}})}]{RANSAC}
Fischler MA and Bolles RC (1981{\natexlab{b}}) Random sample consensus: A
  paradigm for model fitting with applications to image analysis and automated
  cartography.
\newblock \emph{Communications of the ACM} .

\bibitem[{Gelfand et~al.(2005)Gelfand, Mitra, Guibas and
  Pottmann}]{Gelfand:2005aa}
Gelfand N, Mitra N, Guibas L and Pottmann H (2005) {Robust Global
  Registration}.
\newblock In: \emph{Proc. of the Third Eurographics Symposium on Geometry
  Processing}.

\bibitem[{Hernandez et~al.(2016)Hernandez, Bharatheesha, Ko, Gaiser, Tan, van
  Deurzen, de~Vries, Van~Mil, van Egmond, Burger, Morariu, Ju, Gerrmann,
  Ensing, Frankenhuyzen and Wisse}]{hernandez2016team}
Hernandez C, Bharatheesha M, Ko W, Gaiser H, Tan J, van Deurzen K, de~Vries M,
  Van~Mil B, van Egmond J, Burger R, Morariu M, Ju J, Gerrmann X, Ensing R,
  Frankenhuyzen JV and Wisse M (2016) Team delft's robot winner of the amazon
  picking challenge 2016.
\newblock In: \emph{Robot World Cup}. Springer, pp. 613--624.

\bibitem[{Hinterstoisser et~al.(2012)Hinterstoisser, Lepetit, Ilic, Holzer,
  Bradski, Konolige and Navab}]{hinterstoisser2012model}
Hinterstoisser S, Lepetit V, Ilic S, Holzer S, Bradski G, Konolige K and Navab
  N (2012) {Model based training, detection and pose estimation of texture-less
  3D objects in heavily cluttered scenes}.
\newblock In: \emph{{Asian Conference on Computer Vision}}. Springer, pp.
  548--562.

\bibitem[{Hinterstoisser et~al.(2016)Hinterstoisser, Lepetit, Rajkumar and
  Konolige}]{hinterstoisser2016going}
Hinterstoisser S, Lepetit V, Rajkumar N and Konolige K (2016) Going further
  with point pair features.
\newblock In: \emph{European Conference on Computer Vision}. Springer, pp.
  834--848.

\bibitem[{Hodan et~al.(2018)Hodan, Michel, Brachmann, Kehl, Buch, Kraft, Drost,
  Vidal, Ihrke, Zabulis et~al.}]{hodan2018bop}
Hodan T, Michel F, Brachmann E, Kehl W, Buch AG, Kraft D, Drost B, Vidal J,
  Ihrke S, Zabulis X et~al. (2018) Bop: Benchmark for 6d object pose
  estimation.

\bibitem[{Hoda{\v{n}} et~al.(2015)Hoda{\v{n}}, Zabulis, Lourakis,
  Obdr{\v{z}}{\'a}lek and Matas}]{hodavn2015detection}
Hoda{\v{n}} T, Zabulis X, Lourakis M, Obdr{\v{z}}{\'a}lek {\v{S}} and Matas J
  (2015) Detection and fine 3d pose estimation of texture-less objects in rgb-d
  images.
\newblock In: \emph{Intelligent Robots and Systems (IROS), 2015 IEEE/RSJ
  International Conference on}. IEEE, pp. 4421--4428.

\bibitem[{Irani and Raghavan(1996)}]{Irani:1996aa}
Irani S and Raghavan P (1996) {Combinatorial and Experimental Results for
  Randomized Point Matching Algorithms}.
\newblock In: \emph{Proc. of the Symposium on Computational Geometry}. pp.
  68--77.

\bibitem[{Izadi et~al.(2011)Izadi, Kim, Hilliges, Molyneaux, Newcombe, Kohli,
  Shotton, Hodges, Freeman, Davison et~al.}]{izadi2011kinectfusion}
Izadi S, Kim D, Hilliges O, Molyneaux D, Newcombe R, Kohli P, Shotton J, Hodges
  S, Freeman D, Davison A et~al. (2011) Kinectfusion: real-time 3d
  reconstruction and interaction using a moving depth camera.
\newblock In: \emph{Proceedings of the 24th annual ACM symposium on User
  interface software and technology}. ACM, pp. 559--568.

\bibitem[{Johnson and Hebert(1999)}]{johnson1999using}
Johnson AE and Hebert M (1999) Using spin images for efficient object
  recognition in cluttered 3d scenes.
\newblock \emph{IEEE Transactions on pattern analysis and machine intelligence}
  21(5): 433--449.

\bibitem[{Kehl et~al.(2017)Kehl, Manhardt, Tombari, Ilic and
  Navab}]{kehl2017ssd}
Kehl W, Manhardt F, Tombari F, Ilic S and Navab N (2017) Ssd-6d: Making
  rgb-based 3d detection and 6d pose estimation great again.
\newblock In: \emph{IEEE Conference on Computer Vision and Pattern Recognition
  (CVPR)}. pp. 1521--1529.

\bibitem[{Kehl et~al.(2016)Kehl, Milletari, Tombari, Ilic and
  Navab}]{Kehl:2016aa}
Kehl W, Milletari F, Tombari F, Ilic S and Navab N (2016) {Deep Learning of
  Local RGB-D Patches for 3D Object Detection and 6D Pose Estimation}.
\newblock In: \emph{European Conference on Computer Vision (ECCV)}. pp.
  205--220.

\bibitem[{Kim and Medioni(2011)}]{kim20113d}
Kim E and Medioni G (2011) 3d object recognition in range images using
  visibility context.
\newblock In: \emph{Intelligent Robots and Systems (IROS), 2011 IEEE/RSJ
  International Conference on}. IEEE, pp. 3800--3807.

\bibitem[{Kimmel et~al.(2012)Kimmel, Dobson, Littlefield, Krontiris, Marble and
  Bekris}]{kimmel2012pracsys}
Kimmel A, Dobson A, Littlefield Z, Krontiris A, Marble J and Bekris K (2012)
  Pracsys: An extensible architecture for composing motion controllers and
  planners.
\newblock \emph{Int. Conf. on Simulation, Modeling, and Programming for
  Autonomous Robots} : 137--148.

\bibitem[{Kocsis and Szepesv{\'a}ri(2006)}]{kocsis2006bandit}
Kocsis L and Szepesv{\'a}ri C (2006) Bandit based monte-carlo planning.
\newblock In: \emph{ECML}, volume~6. Springer, pp. 282--293.

\bibitem[{Krull et~al.(2015)Krull, Brachmann, Michel, Ying~Yang, Gumhold and
  Rother}]{krull2015learning}
Krull A, Brachmann E, Michel F, Ying~Yang M, Gumhold S and Rother C (2015)
  Learning analysis-by-synthesis for 6d pose estimation in rgb-d images.
\newblock In: \emph{Proceedings of the IEEE International Conference on
  Computer Vision}. pp. 954--962.

\bibitem[{Littlefield et~al.(2015)Littlefield, Krontiris, Kimmel, Dobson, Shome
  and Bekris}]{Littlefield:2015aa}
Littlefield Z, Krontiris A, Kimmel A, Dobson A, Shome R and Bekris KE (2015)
  {An Extensible Software Architecture for Composing Motion and Task Planners}.
\newblock In: \emph{Int. Conf. on Simulation, Modeling and Programming for
  Autonomous Robots (SIMPAR)}.

\bibitem[{Long et~al.(2015)Long, Shelhamer and Darrell}]{shelhamer2016fully}
Long J, Shelhamer E and Darrell T (2015) Fully convolutional networks for
  semantic segmentation.
\newblock In: \emph{IEEE Conf. on Computer Vision and Pattern Recognition}. pp.
  3431--3440.

\bibitem[{Lowe(1999)}]{Lowe:1999aa}
Lowe DG (1999) {Object Recognition from Local Scale-Invariant Features}.
\newblock In: \emph{IEEE International Conference on Computer Vision (ICCV)},
  volume~2. pp. 1150--1157.

\bibitem[{McCormac et~al.(2018)McCormac, Clark, Bloesch, Davison and
  Leutenegger}]{mccormac2018fusion++}
McCormac J, Clark R, Bloesch M, Davison A and Leutenegger S (2018) Fusion++:
  Volumetric object-level slam.
\newblock In: \emph{2018 International Conference on 3D Vision (3DV)}. IEEE,
  pp. 32--41.

\bibitem[{McCormac et~al.(2017)McCormac, Handa, Leutenegger and
  Davison}]{mccormac2017scenenet}
McCormac J, Handa A, Leutenegger S and Davison AJ (2017) Scenenet rgb-d: Can 5m
  synthetic images beat generic imagenet pre-training on indoor segmentation.
\newblock In: \emph{Proceedings of the International Conference on Computer
  Vision (ICCV)}, volume~4.

\bibitem[{Mellado et~al.(2014)Mellado, Aiger and Mitra}]{mellado2014super}
Mellado N, Aiger D and Mitra NJ (2014) {Super4PCS Fast Global Pointcloud
  Registration via Smart Indexing}.
\newblock In: \emph{Computer Graphics Forum}, volume~33. Wiley Online Library,
  pp. 205--215.

\bibitem[{Michel et~al.(2017)Michel, Kirillov, Brachmann, Krull, Gumhold,
  Savchynskyy and Rother}]{michel2017global}
Michel F, Kirillov A, Brachmann E, Krull A, Gumhold S, Savchynskyy B and Rother
  C (2017) Global hypothesis generation for 6d object pose estimation.
\newblock In: \emph{Proceedings of the IEEE Conference on Computer Vision and
  Pattern Recognition}. pp. 462--471.

\bibitem[{Mitash et~al.(2017)Mitash, Bekris and Boularias}]{mitash2017self}
Mitash C, Bekris KE and Boularias A (2017) A self-supervised learning system
  for object detection using physics simulation and multi-view pose estimation.
\newblock In: \emph{Intelligent Robots and Systems (IROS), 2017 IEEE/RSJ
  International Conference on}. IEEE, pp. 545--551.

\bibitem[{Mitash et~al.(2018)Mitash, Boularias and
  Bekris}]{mitash2018improving}
Mitash C, Boularias A and Bekris KE (2018) Improving 6d pose estimation of
  objects in clutter via physics-aware monte carlo tree search.
\newblock In: \emph{2018 IEEE International Conference on Robotics and
  Automation (ICRA)}. IEEE, pp. 1--8.

\bibitem[{Mitra et~al.(2004)Mitra, Gelfand, Pottmann and Guibas}]{Mitra:2004aa}
Mitra N, Gelfand N, Pottmann H and Guibas H (2004) {Registration of Point Cloud
  Data from a Geometric Optimization Perspective}.
\newblock In: \emph{Proc. of the 2004 Eurographics/ACM SIGGRAPH Symposium on
  Geometry Processing}. pp. 22--31.

\bibitem[{Movshovitz-Attias et~al.(2016)Movshovitz-Attias, Kanade and
  Sheikh}]{movshovitz2016useful}
Movshovitz-Attias Y, Kanade T and Sheikh Y (2016) How useful is photo-realistic
  rendering for visual learning?
\newblock In: \emph{ECCV 2016 Workshops}.

\bibitem[{Narayanan and Likhachev(2016)}]{narayanan2016discriminatively}
Narayanan V and Likhachev M (2016) Discriminatively-guided deliberative
  perception for pose estimation of multiple 3d object instances.
\newblock In: \emph{Robotics: Science and Systems}.

\bibitem[{Papazov and Burschka(2010)}]{papazov2010efficient}
Papazov C and Burschka D (2010) An efficient ransac for 3d object recognition
  in noisy and occluded scenes.
\newblock In: \emph{Asian Conference on Computer Vision}. Springer, pp.
  135--148.

\bibitem[{Pavlakos et~al.(2017)Pavlakos, Zhou, Chan, Derpanis and
  Daniilidis}]{Pavlakos:2017aa}
Pavlakos G, Zhou X, Chan A, Derpanis G and Daniilidis K (2017) {6-DOF Object
  Pose from Semantic Keypoints}.
\newblock In: \emph{{IEEE} International Conference on Robotics and Automation
  (ICRA)}.

\bibitem[{Peng et~al.(2015)Peng, Sun, Ali and Saenko}]{peng2015learning}
Peng X, Sun B, Ali K and Saenko K (2015) {Learning deep object detectors from
  3D models}.
\newblock In: \emph{IEEE Intern. Conf. on Computer Vision}.

\bibitem[{Pillai and Leonard(2015)}]{Pillai:2015aa}
Pillai S and Leonard JJ (2015) "monocular slam supported object recognition".
\newblock In: \emph{Robotics: Science and Systems}.

\bibitem[{Redmon et~al.(2016)Redmon, Divvala, Girshick and
  Farhadi}]{redmon2016you}
Redmon J, Divvala S, Girshick R and Farhadi A (2016) You only look once:
  Unified, real-time object detection.
\newblock In: \emph{Proceedings of the IEEE conference on computer vision and
  pattern recognition}. pp. 779--788.

\bibitem[{Ren et~al.(2015)Ren, He, Girshick and Sun}]{ren2015faster}
Ren S, He K, Girshick R and Sun J (2015) Faster r-cnn: Towards real-time object
  detection with region proposal networks.
\newblock In: \emph{Advances in Neural Information Processing Systems}. pp.
  91--99.

\bibitem[{Rennie et~al.(2016)Rennie, Shome, Bekris and
  De~Souza}]{rennie2016dataset}
Rennie C, Shome R, Bekris KE and De~Souza AF (2016) A dataset for improved
  rgbd-based object detection and pose estimation for warehouse pick-and-place.
\newblock \emph{IEEE Robotics and Automation Letters} 1(2): 1179 -- 1185.

\bibitem[{Rothganger et~al.(2006)Rothganger, Lazebnik, Schmid and
  Ponce}]{Rothganger:2006aa}
Rothganger F, Lazebnik S, Schmid C and Ponce J (2006) {3D Object Modeling and
  Recognition using Local Affine-Invariant Image Descriptors and Multi-view
  Spatial Constraints}.
\newblock \emph{International Journal of Computer Vision (IJCV)} 66(3):
  231--259.

\bibitem[{Rusinkiewicz and Levoy(2001)}]{Rusinkiewicz:2001aa}
Rusinkiewicz S and Levoy M (2001) {Efficient Variants of the ICP Algorithm}.
\newblock In: \emph{IEEE Proc. of 3DIM}. pp. 145--152.

\bibitem[{Rusu et~al.(2009)Rusu, Blodow and Beetz}]{rusu2009fast}
Rusu RB, Blodow N and Beetz M (2009) Fast point feature histograms (fpfh) for
  3d registration.
\newblock In: \emph{Robotics and Automation, 2009. ICRA'09. IEEE International
  Conference on}. IEEE, pp. 3212--3217.

\bibitem[{Salas-Moreno et~al.(2013)Salas-Moreno, Newcombe, Strasdat, Kelly and
  Davison}]{salas2013slam++}
Salas-Moreno RF, Newcombe RA, Strasdat H, Kelly PH and Davison AJ (2013)
  Slam++: Simultaneous localisation and mapping at the level of objects.
\newblock In: \emph{Proceedings of the IEEE conference on computer vision and
  pattern recognition}. pp. 1352--1359.

\bibitem[{Segal et~al.(2009)Segal, Haehnel and Thrun}]{Segal:2009aa}
Segal A, Haehnel D and Thrun S (2009) {Generalized-ICP}.
\newblock In: \emph{Robotics: Science and Systems}, volume~2. p.~4.

\bibitem[{Shrivastava et~al.(2017)Shrivastava, Pfister, Tuzel, Susskind, Wang
  and Webb}]{shrivastava2017learning}
Shrivastava A, Pfister T, Tuzel O, Susskind J, Wang W and Webb R (2017)
  Learning from simulated and unsupervised images through adversarial training.
\newblock In: \emph{The IEEE Conference on Computer Vision and Pattern
  Recognition (CVPR)}, volume~3. p.~6.

\bibitem[{Simonyan and Zisserman(2015)}]{simonyan2014very}
Simonyan K and Zisserman A (2015) Very deep convolutional networks for
  large-scale image recognition.
\newblock \emph{International Conference on Learning Representations (ICLR)} .

\bibitem[{Singh et~al.(2014)Singh, Sha, Narayan, Achim and
  Abbeel}]{singh2014bigbird}
Singh A, Sha J, Narayan KS, Achim T and Abbeel P (2014) Bigbird: A large-scale
  3d database of object instances.
\newblock In: \emph{IEEE International Conference on Robotics and Automation
  (ICRA)}. IEEE.

\bibitem[{Srivatsan et~al.(2017)Srivatsan, Vagdargi and
  Choset}]{Srivatsan:2017aa}
Srivatsan RA, Vagdargi P and Choset H (2017) {Sparse Point Registration}.
\newblock In: \emph{International Symposium on Robotics Research (ISRR)}.

\bibitem[{Stein and Roy(2018)}]{stein2018genesis}
Stein GJ and Roy N (2018) Genesis-rt: Generating synthetic images for training
  secondary real-world tasks.
\newblock In: \emph{2018 IEEE International Conference on Robotics and
  Automation (ICRA)}. IEEE, pp. 7151--7158.

\bibitem[{Su et~al.(2015)Su, Qi, Li and Guibas}]{su2015render}
Su H, Qi CR, Li Y and Guibas LJ (2015) {Render for CNN: Viewpoint estimation in
  images using CNNs trained with rendered 3d model views}.
\newblock In: \emph{IEEE Intern. Conf. on Computer Vision}.

\bibitem[{Sun and Saenko(2014)}]{SunVirtual}
Sun B and Saenko K (2014) From virtual to reality: Fast adaptation of virtual
  object detectors to real domains.
\newblock In: \emph{British Machine Vision Conf.}

\bibitem[{Tejani et~al.(2014)Tejani, Tang, Kouskouridas and
  Kim}]{tejani2014latent}
Tejani A, Tang D, Kouskouridas R and Kim TK (2014) Latent-class hough forests
  for 3d object detection and pose estimation.
\newblock In: \emph{European Conference on Computer Vision}. Springer, pp.
  462--477.

\bibitem[{Thrun et~al.(2005)Thrun, Burgard and Fox}]{thrun2005probabilistic}
Thrun S, Burgard W and Fox D (2005) \emph{Probabilistic robotics}.
\newblock MIT press.

\bibitem[{Tobin et~al.(2017)Tobin, Fong, Ray, Schneider, Zaremba and
  Abbeel}]{tobin2017domain}
Tobin J, Fong R, Ray A, Schneider J, Zaremba W and Abbeel P (2017) Domain
  randomization for transferring deep neural networks from simulation to the
  real world.
\newblock In: \emph{Intelligent Robots and Systems (IROS), 2017 IEEE/RSJ
  International Conference on}. IEEE, pp. 23--30.

\bibitem[{Tombari and Di~Stefano(2010)}]{tombari2010object}
Tombari F and Di~Stefano L (2010) Object recognition in 3d scenes with
  occlusions and clutter by hough voting.
\newblock In: \emph{Image and Video Technology (PSIVT), 2010 Fourth Pacific-Rim
  Symposium on}. IEEE, pp. 349--355.

\bibitem[{Tombari et~al.(2010)Tombari, Salti and
  Di~Stefano}]{tombari2010unique}
Tombari F, Salti S and Di~Stefano L (2010) Unique signatures of histograms for
  local surface description.
\newblock In: \emph{European conference on computer vision}. Springer, pp.
  356--369.

\bibitem[{Vidal et~al.(2018)Vidal, Lin and Mart{\'\i}}]{vidal20186d}
Vidal J, Lin CY and Mart{\'\i} R (2018) 6d pose estimation using an improved
  method based on point pair features.
\newblock In: \emph{2018 4th International Conference on Control, Automation
  and Robotics (ICCAR)}. IEEE, pp. 405--409.

\bibitem[{Wohlhart and Lepetit(2015)}]{Wohlhart:2015aa}
Wohlhart P and Lepetit V (2015) {Learning Descriptors for Object Recognition
  and 3D Pose Estimation}.
\newblock In: \emph{Conference on Computer Vision and Pattern Recognition
  (CVPR)}.

\bibitem[{Xiang et~al.(2018)Xiang, Schmidt, Narayanan and
  Fox}]{xiang2017posecnn}
Xiang Y, Schmidt T, Narayanan V and Fox D (2018) Posecnn: A convolutional
  neural network for 6d object pose estimation in cluttered scenes.
\newblock \emph{Robotics: Science and Systems (RSS)} .

\bibitem[{Zeng et~al.(2017)Zeng, Yu, Song, Suo, Walker~Jr, Rodriguez and
  Xiao}]{Princeton}
Zeng A, Yu KT, Song S, Suo D, Walker~Jr E, Rodriguez A and Xiao J (2017)
  Multi-view self-supervised deep learning for 6d pose estimation in the amazon
  picking challenge.
\newblock In: \emph{{IEEE} International Conference on Robotics and Automation
  (ICRA)}.

\bibitem[{Zhang et~al.(2007)Zhang, Kim and Manocha}]{zhang2007c}
Zhang L, Kim YJ and Manocha D (2007) C-dist: efficient distance computation for
  rigid and articulated models in configuration space.
\newblock In: \emph{Proceedings of the 2007 ACM symposium on Solid and physical
  modeling}. ACM, pp. 159--169.

\bibitem[{Zhou et~al.(2016)Zhou, Park and Koltun}]{koltun}
Zhou QY, Park J and Koltun K (2016) {Fast Global Registration}.
\newblock \emph{European Conference on Computer Vision} .

\bibitem[{Zhu et~al.(2017)Zhu, Park, Isola and Efros}]{zhu2017unpaired}
Zhu JY, Park T, Isola P and Efros AA (2017) Unpaired image-to-image translation
  using cycle-consistent adversarial networks.
\newblock In: \emph{Proceedings of the International Conference on Computer
  Vision (ICCV)}.

\end{thebibliography}

\end{document}